%% file: main.tex
\documentclass[sigconf]{acmart}
\AtBeginDocument{%
  }

\usepackage{thmtools}
\usepackage{enumitem}
\usepackage{amsmath}
\usepackage{pifont}
\usepackage{subcaption}
\usepackage{cleveref}
\usepackage[ruled,linesnumbered]{algorithm2e}
\usepackage{multirow}

\DeclareMathOperator*{\graph}{\mathcal{G}}
\newcommand{\rv}[1]{\boldsymbol{\mathrm{#1}}} %
\newcommand{\best}[1]{\textbf{#1}} %
\newcommand{\secbest}[1]{\underline{#1}} %

\AtEndPreamble{%
 \theoremstyle{acmplain}
 \newtheorem{assumption}[theorem]{Assumption}}

\AtEndPreamble{%
 \theoremstyle{acmplain}
 \newtheorem{problem}[theorem]{PROBLEM}}

\newcommand{\name}{\textbf{GOODFormer}}
\newcommand{\disentangler}{entropy-guided invariant subgraph disentangler}
\newcommand{\encoder}{evolving subgraph positional and structural encoder}
\newcommand{\disentanglertitle}{Entropy-guided Invariant Subgraph Disentangler}
\newcommand{\encodertitle}{Evolving Subgraph Positional and Structural Encoder}

\copyrightyear{2026}
\acmYear{2026}
\setcopyright{cc}
\setcctype{by-nc-nd}
\acmConference[KDD '26]{Proceedings of the 32nd ACM SIGKDD Conference on Knowledge Discovery and Data Mining V.1}{August 09--13, 2026}{Jeju Island, Republic of Korea}
\acmBooktitle{Proceedings of the 32nd ACM SIGKDD Conference on Knowledge Discovery and Data Mining V.1 (KDD '26), August 09--13, 2026, Jeju Island, Republic of Korea}
\acmPrice{}
\acmDOI{10.1145/3770854.3780247}
\acmISBN{979-8-4007-2258-5/2026/08}

\acmSubmissionID{558}

\begin{document}

\title{Invariant Graph Transformer for Out-of-Distribution Generalization}

\author{Tianyin Liao}
\affiliation{
  \institution{Nankai University}
  \city{Binhai New Area}
  \state{Tianjin}
  \country{China}
}
\email{1120230329@mail.nankai.edu.cn}

\author{Ziwei Zhang}
\authornote{Corresponding author.}
\affiliation{
  \institution{Beihang University}
  \city{Haidian District}
  \state{Beijing}
  \country{China}
}
\email{zwzhang@buaa.edu.cn}

\author{Yufei Sun}
\affiliation{
  \institution{Nankai University}
  \city{Binhai New Area}
  \state{Tianjin}
  \country{China}
}
\email{yufei_sun@sina.com}

\author{Chunyu Hu}
\affiliation{
  \institution{Nankai University}
  \city{Binhai New Area}
  \state{Tianjin}
  \country{China}
}
\email{huchunyu@mail.nankai.edu.cn}

\author{Jianxin Li}
\affiliation{
  \institution{Beihang University}
  \city{Haidian District}
  \state{Beijing}
  \country{China}
}
\email{lijx@buaa.edu.cn}

\renewcommand{\shortauthors}{Tianyin Liao, Ziwei Zhang, Yufei Sun, Chunyu Hu, \& Jianxin Li}

\begin{abstract}
Graph Transformers (GTs) have demonstrated great effectiveness across various graph analytical tasks. However, the existing GTs focus on training and testing graph data originated from the same distribution, but fail to generalize under distribution shifts. Graph invariant learning, aiming to capture generalizable graph structural patterns with labels under distribution shifts, is potentially a promising solution, but how to design attention mechanisms and positional and structural encodings (PSEs) based on graph invariant learning principles remains challenging. To solve these challenges, we introduce \underline{G}raph \underline{O}ut-\underline{O}f-\underline{D}istribution generalized Trans\underline{former} (\name), aiming to learn generalized graph representations by capturing invariant relationships between predictive graph structures and labels through jointly optimizing three modules. Specifically, we first develop a GT-based \disentangler~to separate invariant and variant subgraphs while preserving the sharpness of the attention function. Next, we design an \encoder~to effectively and efficiently capture the encoding information of dynamically changing subgraphs during training. Finally, we propose an invariant learning module utilizing subgraph node representations and encodings to derive generalizable graph representations that can  to unseen graphs. We also provide theoretical justifications for our method. Extensive experiments on benchmark datasets demonstrate the superiority of our method over state-of-the-art baselines under distribution shifts.
\end{abstract}



\maketitle
\newcommand\kddavailabilityurl{https://doi.org/10.5281/zenodo.18058301}
\ifdefempty{\kddavailabilityurl}{}{
\begingroup\small\noindent\raggedright\textbf{Resource Availability:}\\
The source code of this paper has been made publicly available at \url{\kddavailabilityurl} and can also be accessed via the project repository: \url{https://github.com/Lowy999/GOODFormer}.
\endgroup
}

\input{sections/1-Intro}
\input{sections/2-Form}
\input{sections/3-Method}
\input{sections/4-Exp}

\input{sections/5-Related}
\input{sections/6-Con}

\begin{acks}
This work was supported in part by the National Natural Science Foundation of China (No. 62472018, 92570113) and the Fundamental Research Funds for the Central Universities JK2024-07.
\end{acks}

\bibliographystyle{ACM-Reference-Format}
\balance
\bibliography{base}

\appendix
\input{appendices/Impact}
\input{appendices/Limitation}
\input{appendices/Notation}
\input{appendices/Theory}
\input{appendices/ExpDetail}
\input{appendices/MoreExp}

\end{document}

%% file: sections/1-Intro.tex
\section{Introduction}\label{sec:intro}
Graph learning has become a critical and rapidly evolving field~\cite{zhang2020deep,wu2020comprehensive}, achieving impressive outcomes across various domains, such as modeling molecules~\cite{you2018graph}, social networks~\cite{fan2019graph}, and knowledge graphs~\cite{battaglia2018relational}. While much of the focus has been on graph neural networks (GNNs)~\cite{kipf2022semi,velivckovic2018graph,hamilton2017inductive,xu2018powerful} that rely on local message passing~\cite{gilmer2017neural}, graph Transformers (GTs) have recently gained significant attention~\cite{shehzad2024graph,min2022transformer}. In short, GTs adapt the Transformer architecture~\cite{waswani2017attention}, which has shown success in sequential data tasks like natural language processing~\cite{tunstall2022natural}, to effectively capture graph structures through various modifications. One notable advantage of GTs is that they allow nodes to interact with all other nodes in a graph, enabling the modeling of long-range dependencies~\cite{ying2021transformers}. This approach addresses several limitations of GNNs, such as oversmoothing~\cite{chen2020measuring}, over-squashing~\cite{toppingunderstanding}, and limited expressiveness~\cite{sato2020survey}. The potential of GTs has led to a surge in innovative model designs in recent years~\cite{min2022transformer,kreuzer2021rethinking,rampavsek2022recipe,shirzad2023exphormer,dwivedi2020generalization,chen2022structure}.

However, despite their success, current GTs often depend on the assumption that training and testing graph data come from the same distribution. In practice, distribution shifts frequently occur due to the unpredictable nature of real-world data generation~\cite{liu2021towards}. Without careful designs, the existing models struggle with generalizing to out-of-distribution (OOD) graph data as they tend to exploit statistical correlations between graph data and label, which could easily be variable under distribution shifts~\cite{li2022out}. Failing to address the OOD generalization problem  will make graph machine learning models unuseful, especially for high-stakes graph applications like medical diagnosis, financial analysis, molecular prediction, etc.
Therefore, how to develop GTs that can handle spurious correlations and learn graph representations capable of OOD generalization is a crucial but largely overlooked problem.

In this work, we tackle the OOD generalization problem of GTs using the invariant learning principle. Invariant learning aims to capture stable relationships between features and labels across varying distributions, while ignoring spurious correlations. Some methods are proposed to tackle the OOD generalization problem of GNNs using invariant learning~\cite{wudiscovering,li2022learning}. Typically, these methods combine invariant learning with the message-passing mechanism of GNNs to discover invariant subgraphs. However, directly applying invariant graph learning to GTs faces the following two challenges:

First, for GTs, the self-attention function plays a central role in distinguishing invariant and variant subgraphs. However, when dealing with distribution shifts, the classical self-attention cannot guarantee sharpness~\cite{velivckovic2024softmax}. Lack of sharpness can lead to incomplete disentanglement between invariant and variant subgraphs. For example, if the number of variant subgraph nodes in the test data greatly exceeds that in the training data, the attention coefficients assigned to variant subgraph nodes will disperse, which reduces the margin in attention coefficients between invariant and variant subgraph nodes. We provide a formal theoretical analysis for this problem in Theorem~\ref{theorem: softmax disperse}.

Second, positional and structural encoding (PSE) is critical to empower GTs capably of capturing valuable information in graph data and enhance its expressiveness~\cite{li2024improves}. However, in invariant graph learning methods, the invariant subgraph of the same input graph dynamically evolves as the model is optimized. If we directly compute hand-crafted PSEs for these evolving subgraphs, the computational complexity is prohibitive. How to design an encoding mechanism that can efficiently generate PSEs for evolving subgraphs while maintaining the theoretical expressiveness remains unexplored.

To address these challenges, we propose \name, Graph Out-Of-Distribution Generalized Transformer
. Our proposed method handles the distribution shifts of graphs by capturing invariant subgraphs and learning OOD generalized graph representations while resolving the aforementioned two challenges. Specifically, as shown in Figure \ref{fig:framework}, our proposed method achieves this goal by jointly optimizing three mutually reinforcing modules. First, we design a GT-based \disentangler~to separate invariant and variant subgraphs using attention coefficients. To maintain the sharpness of the attention function, we propose entropy-guided sharp attention, comprising an entropy-guided regularizer and a lightweight temperature-based test-time training method. Next, we introduce an \encoder~to learn PSEs of dynamically changing subgraphs guided by the hand-crafted PSEs. In this way, our method can efficiently generate PSEs while maintaining the expressiveness. Finally, we propose an invariant learning module to learn graph representations that generalize to unknown test graphs through tailored objective functions. We also provide theoretical justifications for our proposed method in maintaining the sharpness of the attention function.
We conduct extensive experiments on both synthetic and real-world graph benchmarks for the graph classification task. The results show that \name~consistently and significantly outperforms various state-of-the-art baselines on unseen OOD testing graphs. Ablation studies and hyper-parameter analysis further demonstrate the designs of our proposed method. 

Our contributions are summarized as follows:
\begin{itemize}[leftmargin=0.30cm]
\item We propose a novel \name~model to enhance the OOD generalization capabilities of GTs. To our knowledge, this is the first study for GTs under distribution shifts.
\item We develop an \disentangler~to effectively separate invariant and variant subgraphs for GTs, while maintaining the sharpness of the attention function with theoretical guarantees.
\item We further design an \encoder~to capture the positional and structural information of dynamically evolving subgraphs during training, maintaining efficiency as well as expressiveness.
\item  Extensive empirical results demonstrate the effectiveness of \name~across various synthetic and real-world benchmark graph datasets under distribution shifts.
\end{itemize}

The rest of the paper is organized as follows. In Section~\ref{sec:notations}, we introduce notations and the problem formulation. In Section~\ref{sec:method}, we introduce our proposed method in detail. In Section~\ref{sec:exp}, we report experimental results, together with ablation studies and hyper-parameter analyses. We review related works in Section~\ref{sec:related} and conclude the paper in Section~\ref{sec:conclusion}.

%% file: sections/2-Form.tex
\section{Notations and Problem Formulation}\label{sec:notations}
Let $\mathbb{G}$ and $\mathbb{Y}$ represent the graph and label space and $\rv{G}$ and $\rv{Y}$ denote the random variables corresponding to the graph and label, respectively.  In this paper, we mainly focus on the graph classification task. Consider a graph dataset $\graph=\{(G_i,Y_i)\}^{N}_{i=1}$, where $G_{i}\in \mathbb{G}$ and $Y_{i}\in \mathbb{Y}$. Concretely, for a graph instance $G=(\mathcal{V}, \mathcal{E})$ with the node set $\mathcal{V}$ and the edge set $\mathcal{E}$, its adjacency matrix is $\boldsymbol{A} \in \{0,1\}^{\lvert\mathcal{V}\rvert\times\lvert\mathcal{V}\rvert}$, where $\boldsymbol{A}_{ij}=1$ denotes that there exists an edge from node $i$ to node $j$, and $\boldsymbol{A}_{ij}=0$ otherwise. We formulate the generalization problem under distribution shifts on graphs as:
\begin{problem}
Given a training graph dataset $\mathcal{G}_{\mathrm{train}}$, the task is to learn a graph predictor $h(\cdot):\mathbb{G} \rightarrow \mathbb{Y}$ that achieves satisfactory performance on an unseen test graph dataset $\mathcal{G}_{\mathrm{test}}$ where there exist distribution shifts, i.e., $\rm{P}(\mathcal{G}_{\mathrm{train}}) \neq \rm{P}(\rm{\mathcal{G}_{\mathrm{test}}})$.
\end{problem}
Note that the distribution shifts make standard empirical risk minimization (ERM), e.g., $\arg \min_h \sum_{i=1}^N \mathcal{L}(h(G_i),Y_i)$, where $\mathcal{L}(\cdot)$ is a loss function, such as the cross-entropy loss, unable to obtain satisfactory performance.
 
Inspired by causal theory~\cite{pearl2009causality}, we assume that $\rv{G}$ contains an invariant subgraph $\rv{G}_C \subset \rv{G}$, whose relationship with the label remains invariant across different environments. We refer to the rest of the graph as the variant subgraph, denoted by $\rv{G}_S = \rv{G} \backslash \rv{G}_C$. There exists a function $f_Y:\rv{G}_C\rightarrow \rv{Y}$, where the "oracle rationale" $\rv{G}_C$ satisfies:
\begin{equation}\label{eq:GTObj}
  Y=f_Y(\rv{G}_C),\quad \rv{Y} \perp \rv{G}_S \mid \rv{G}_C,
\end{equation}
where $\rv{Y} \perp \rv{G}_S \mid \rv{G}_C$ indicates that $\rv{G}_C$ shields $\rv{Y}$ from the influence of $\rv{G}_S$, making the causal relationship $\rv{G}_C \rightarrow \rv{Y}$ invariant across different $\rv{G}_S$.
Generally, during training, only the pairs of input $\rv{G}$ and label $\rv{Y}$ are observed, while neither the oracle rationale $\rv{G}_C$ nor the oracle structural equation model $f_Y$ is accessible. 

In this paper, we focus on GTs as the predictor, which is divided into two components, i.e., $h=h_{\hat{Y}} \circ \Phi$. Specifically, a GT-based disentangler $\Phi:\rv{G}\rightarrow \rv{G}_{\tilde{C}}$ is utilized to identify the potential rationale $\rv{G}_{\tilde{C}}$ from $\rv{G}$, and another GT-based predictor $h_{\hat{Y}}:\rv{G}_{\tilde{C}}\rightarrow\hat{\rv{Y}}$ generates the prediction from $\rv{G}_{\tilde{C}}$ to approach $\rv{Y}$ and incorporates a positional and structural encoder to enhance the expressiveness of the GT. To optimize these components, we formulate the task of invariant rationalization as:
\begin{equation}
  \min \nolimits_{h_{\hat{Y}}, \Phi} \mathcal{L}(h_{\hat{Y}} (\Phi(\rv{G})),\rv{Y}),\quad s.t. \ \rv{Y} \perp \rv{G}_{\tilde{S}}\mid \rv{G}_{\tilde{C}},
\end{equation}
where $\rv{G}_{\tilde{S}}=\rv{G} \backslash \rv{G}_{\tilde{C}}$ is the complement of $\rv{G}_{\tilde{C}}$.

%% file: sections/3-Method.tex
\section{Method}\label{sec:method}

\begin{figure*}
    \centering
    \includegraphics[width=\textwidth]{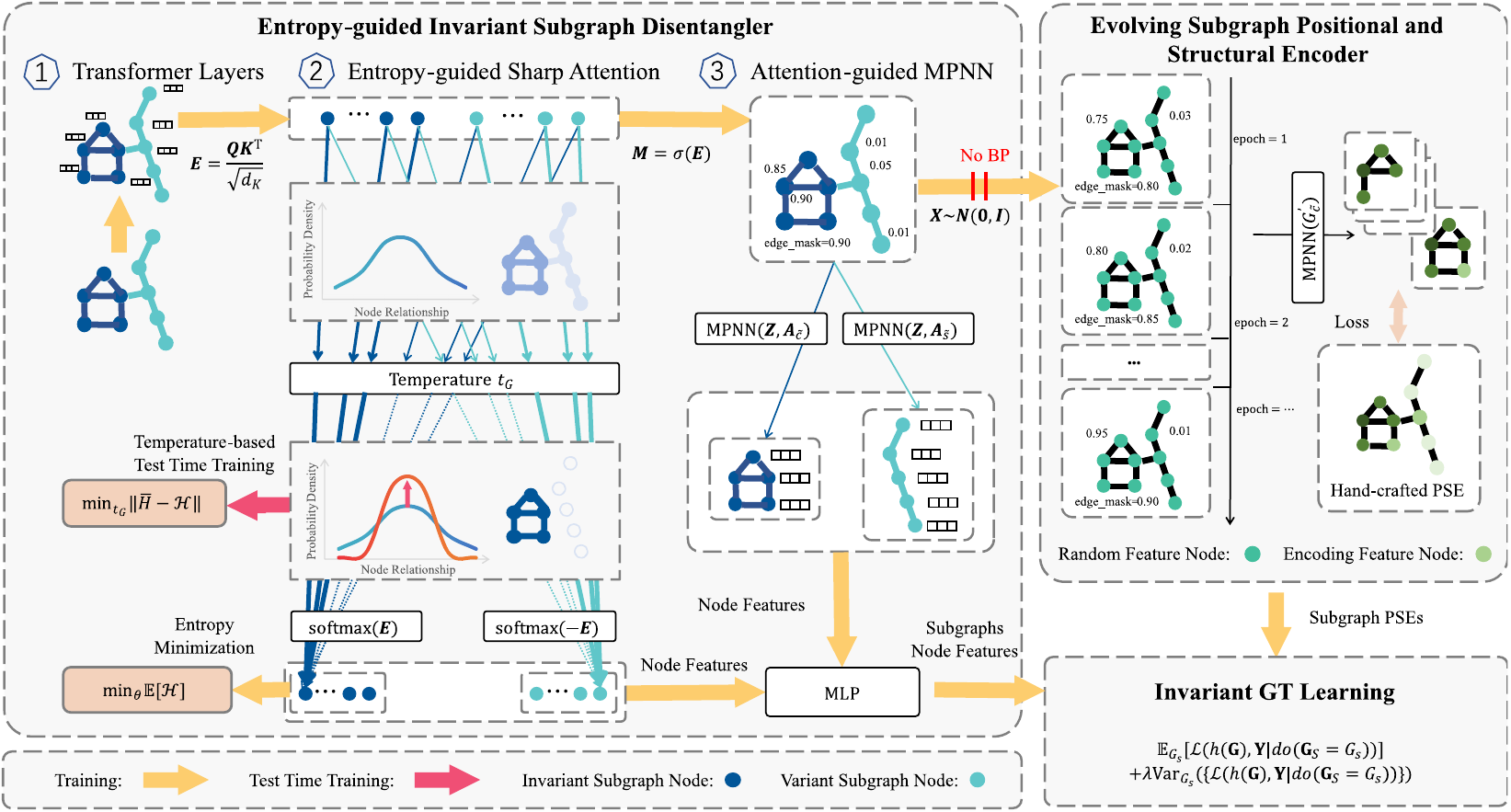}
    \caption{The framework of \name, which jointly optimizes three modules: (1) The \disentangler~utilizes Transformer layers, entropy-guided sharp attention and the attention-guided MPNN to separate invariant and variant subgraphs. (2) The \encoder~captures valuable information of dynamically changing subgraphs during training while maintaining expressiveness. (3) The invariant GT learning module optimizes tailored objective functions to derive graph representations for generalizing to unseen test graphs.}
    \label{fig:framework}
\end{figure*}

In this section, we introduce our proposed \name. The framework is illustrated in Figure \ref{fig:framework}. We begin with an introduction to the \disentangler~module. Then, we present the evolving subgraph positional and structural
encoder module. Next, we describe the invariant learning module, which is designed to derive invariant graph representations from the subgraph features and encodings. Lastly, we provide some discussions of our proposed method. 

\subsection{\disentanglertitle}
The disentangler aims to split the input graph instance $G$ into two subgraphs: invariant subgraph $G_{\tilde{c}}$ and variant subgraph $G_{\tilde{s}}$. Specifically, given an input graph instance, the disentangler first calculates two complementary attentions using the softmax function as:
\begin{equation}\label{eq:complementatt}
\begin{aligned}
\boldsymbol{Z}_{\mathrm{Attn},{\tilde{c}}}=\mathrm{Softmax}(\boldsymbol{E})\boldsymbol{Z}\boldsymbol{W}_{V},
\boldsymbol{Z}_{\mathrm{Attn},{\tilde{s}}}=\mathrm{Softmax}(-\boldsymbol{E})\boldsymbol{Z}\boldsymbol{W}_{V}.
\end{aligned}
\end{equation}
where 
$\boldsymbol{Z}_{\mathrm{Attn},{\tilde{c}}}$ and $\boldsymbol{Z}_{\mathrm{Attn},{\tilde{s}}}$ are attention outputs for the invariant and variant subgraph, respectively, 
$\boldsymbol{Z}\in\mathbb{R}^{\lvert\mathcal{V}\rvert\times d}$ are $d$-dimensional intermediate node representations calculated by graph Transformer layers $\mathrm{GT}^M(\cdot)$ and $\boldsymbol{E}$ are intermediate attention scores calculated as $\boldsymbol{E}=\frac{\boldsymbol{Z}\boldsymbol{W}_{Q}(\boldsymbol{Z}\boldsymbol{W}_{K})^\top}{\sqrt{d_{K}}}$, $\boldsymbol{W}_{Q}$, $\boldsymbol{W}_{K}$ and $\boldsymbol{W}_{V}$ are three projection matrices, and  $d_{K}$ is the hidden dimension. In short, these two attentions can distinguish invariant and variant subgraphs due to the  complementarity design, i.e., if $(i,j) \in G_{\tilde{c}}$, then $\boldsymbol{E}_{(i,j)}>0$, and if $(i,j) \in G_{\tilde{s}}$, then $\boldsymbol{E}_{(i,j)}<0$, where $i$ and $j$ represent the indices of the nodes. The $\mathrm{softmax}$ function is used to highlight the node relationships $(i,j)$ that belong to $G_{\tilde{c}}$, while minimizing the scores of those node relationships $(i,j)$ that belong to $G_{\tilde{s}}$ as much as possible, ideally approaching zero. Compared to other invariant subgraphs extraction methods such as Top-K pooling~\cite{zhang2018end}, softmax offers more flexibility and is gradient-friendly since it avoids any discrete operation. Besides, softmax is more adaptive at distinguishing invariant and variant subgraphs compared to the sigmoid function, which exhibits very small gradients when input values approach 0 or 1.  

After obtaining the attentions, we further need to obtain the explicit structure of $G_{\tilde{c}}$ and $G_{\tilde{s}}$. Specifically, we use a soft mask matrix $\boldsymbol{M}$ from the attention logits $\boldsymbol{E}$, and then obtain the adjacency matrix of the invariant and variant subgraphs as follows:
\begin{equation}
\boldsymbol{M}=\sigma(\boldsymbol{E}),\boldsymbol{A}_{\tilde{c}}=\boldsymbol{M}\odot\boldsymbol{A},\boldsymbol{A}_{\tilde{s}}=(1-\boldsymbol{M})\odot\boldsymbol{A},
\label{eq:subgraphadjacencymatrix}
\end{equation}
where $\sigma(\cdot)$ is the sigmoid function, $\boldsymbol{A}_{\tilde{c}}$ and $\boldsymbol{A}_{\tilde{s}}$ denote the adjacency matrix of $G_{\tilde{c}}$ and $G_{\tilde{s}}$, respectively, $\odot$ denotes the element-wise matrix multiplication.

Now that we have the structure of the invariant and variant subgraphs, we adopt an attention-guided MPNN to learn more comprehensive representations of nodes. Specifically, the MPNN layers utilize the adjacency matrices $\boldsymbol{A}_{\tilde{c}}$ and $\boldsymbol{A}_{\tilde{s}}$ to compute the subgraph node representations $\boldsymbol{Z}_{\mathrm{MPNN},{\tilde{c}}}$ and $\boldsymbol{Z}_{\mathrm{MPNN},{\tilde{s}}}$, respectively:
\begin{equation}
\begin{aligned}
\boldsymbol{Z}_{\mathrm{MPNN},{\tilde{c}}}=\mathrm{MPNN} (\boldsymbol{Z},\boldsymbol{A}_{\tilde{c}}), \boldsymbol{Z}_{\mathrm{MPNN},{\tilde{s}}}=\mathrm{MPNN}(\boldsymbol{Z},\boldsymbol{A}_{\tilde{s}}),
\label{eq:mpnnoutput}
\end{aligned}
\end{equation}
Finally, we make our method a hybrid architecture combining MPNN layers and Transformer layers by aggregating the representation of MPNN layers with self-attention layers:
\begin{equation}
\begin{aligned}
\boldsymbol{Z}_{\tilde{c}}=\mathrm{MLP}(\boldsymbol{Z}_{\mathrm{Attn},{\tilde{c}}}+\boldsymbol{Z}_{\mathrm{MPNN},{\tilde{c}}}), \\\boldsymbol{Z}_{\tilde{s}}=\mathrm{MLP}(\boldsymbol{Z}_{\mathrm{Attn},{\tilde{s}}}+\boldsymbol{Z}_{\mathrm{MPNN},{\tilde{s}}}),
\end{aligned}
\label{eq:mlpaggregate}
\end{equation}
where $\mathrm{MLP}$ is composed of fully connected layers.

So far, we have introduced an invariant subgraph disentangler to distinguish invariant and variant subgraphs. However, we find that directly optimizing the disentangler encounters difficulty in establishing clear decision boundaries. Specifically, when dealing with graph data under distribution shifts, the attention function in Eq.~\eqref{eq:complementatt} cannot guarantee sharpness between invariant and variant subgraphs, i.e., there $\exists G \in \mathbb{G}$ that satisfies
\begin{equation}
\mathrm{Softmax}(\boldsymbol{E})_{i,j} - \mathrm{Softmax}(\boldsymbol{-E})_{i,j} < \epsilon,
\end{equation}
where $\mathrm{Softmax}(\boldsymbol{E})_{i,j}$ and $\mathrm{Softmax}(\boldsymbol{-E})_{i,j}$ corresponds to the attention score of node relationship $(i,j)$ in $G_c$ and $G_s$ in Eq.~\eqref{eq:complementatt}, respectively, and $\epsilon$ is a small positive number that ensures the disentangler can correctly distinguish invariant and variant subgraphs.

More formally, we can prove the following theorem: 
\begin{restatable}{theorem}{softmaxdisperse}
Suppose $\rv{G}_S\rightarrow \rv{G}_C$ does not exist, $\mathcal{L}(\cdot)$ is a loss function and there exists one and only one non-trival subgraph $\rv{G}_C$. Under these conditions, any model comprising Eq.~\eqref{eq:complementatt} cannot guarantee  the following intervention-based invariant learning principle: (1) minimizes $\mathbb{E}_{G_s}[\mathcal{L}(h(\rv{G}),\rv{Y}|do(\rv{G}_S=G_s))]$, and simultaneously (2) minimizes $\mathrm{Var}_{G_s}(\{\mathcal{L}(h(\rv{G}),\rv{Y}|do(\rv{G}_S=G_s))\})$,
where $do(\rv{G}_S=G_s)$ denotes the do-calculus operation that removes every link from the parents to the variable $\rv{G}_S$ and fixes $\rv{G}_S$ to the specific value $G_s$.
\label{theorem: softmax disperse}
\end{restatable}
The proof is provided in Appendix~\ref{subsec:proof}. The theorem shows that due to the limitations of the plain attention function, the disentangler cannot guarantee satisfying the principle upheld by the oracle $f_Y$ (given $\rv{G}_C$), which restricts the OOD generalization capability. 
This limitation hampers the OOD generalization capability of our proposed disentangler under the invariant learning principle, which has shown effective within GNN architectures.

To address this potential issue, we propose an entropy-guided regularizer and temperature-based test time training strategy to transform plain self-attention into entropy-guided sharp attention. Assume that the scores of node relationships $\mathrm{Softmax}(\boldsymbol{E})_{(i,j)}$ within the same subgraphs have small differences, and the scores between the invariant subgraph and variant subgraph have larger disparities. We can use entropy to measure the score differences between the node relationships in the invariant and variant subgraphs. A lower entropy value indicates a larger score disparity, making the attention function sharper. Therefore, during the training phase, we minimize the following entropy-guided regularizer entropy of the softmax output to sharpen the attention function:
\begin{equation}
\mathcal{L}_E=\mathbb{E}_{E} [\mathcal{H}(\mathrm{Softmax}(\boldsymbol{E}) )],
\end{equation}
where $\mathcal{H}(\cdot)$ is the entropy function and $\boldsymbol{E}$ are attention scores. Despite minimizing the above entropy can lead to sharp attention during training, the attention scores for testing graphs are not guaranteed to be sharp since the graph data distribution may change drastically. To tackle this issue, we adopt a simply-yet-effective test time training strategy to obtain desirable test-time attention. Specifically, consider a temperature-based softmax function as: \begin{equation}\label{eq:tematt}
    \mathrm{Softmax}_{t}(e_i)=\frac{\mathrm{exp}(e_i/t)}{\sum_{k}\mathrm{exp}(e_k/t)},
\end{equation}
where $t$ is the temperature. It is easy to see that the plain softmax function in Eq.~\eqref{eq:complementatt} is equal to $t=1$. Using this function, we can reduce the temperature to decrease the entropy while maintaining the relative rank of the attention scores between pairs of nodes. Formally, we have the following theorem:
\begin{restatable}{theorem}{temperature}
Consider a vector $\boldsymbol{e} \in \mathbb{R}^n$, where $n>1$, comprising $n$ logits in the temperature-based softmax function Eq.~\eqref{eq:tematt} with $t > 0$. Let $H=-\sum_{i}p_{i}\log p_i, p_{i}=\mathrm{Softmax}_{t}(e_i)$ denote the Shannon entropy of the temperature-based softmax function. As $t$ decreases, $H$ monotonically decreases.
\end{restatable}
The proof is provided in Appendix~\ref{subsec:proof}. The theorem shows that we can adjust the entropy by changing the temperature. Therefore, during the testing stage, we optimize the temperature $t_G$ for each graph $G$. To prevent the model from being overly confident due to an excessively low temperature, we calculate the average entropy $\overline{H}$ during the training phase to guide the test-time training. The objective is to minimize the absolute difference between $\overline{H}$ and the entropy on test data, i.e.,:
\begin{equation}
\min \nolimits_{t_G}\rvert\overline{H}-\mathcal{H}(\mathrm{Softmax}_{t_G}(\boldsymbol{E}_G))\rvert.
\label{eq:testtimetraining}
\end{equation}
Combining the entropy-guided regularizer and temperature-based test time training, our disentangler can effectively distinguish invariant and variant subgraphs for subsequent components.

\subsection{\encodertitle}
After obtaining the invariant subgraph using the disentangler, next we need to learn expressive graph representations. We propose an encoder designed to capture the positional and structural information of graphs to enhance model expressivity, which in turn leads to improved generalization. The existing GTs usually adopt hand-crafted encoding methods during the pre-processing phase due to their high computational complexity~\cite{shehzad2024graph,hoang2024survey}. However, in our proposed method, the invariant subgraphs are dynamically evolving as the disentangler is optimized during training, making computing the encoding of subgraphs during pre-processing infeasible. Therefore, our goal is to efficiently generate PSEs for dynamically changing subgraphs while avoiding spurious correlations. 

Specifically, we propose a learnable MPNN-based subgraph encoder that uses $\boldsymbol{A}_{\tilde{c}}$ and $\boldsymbol{A}_{\tilde{s}}$ from the attention-guided MPNN to generate the positional and structural information of $G_{\tilde{c}}$ and $G_{\tilde{s}}$, respectively.
A key challenge is to empower the model to learn expressive PSEs, such as the Laplacian Positional Encoding~\cite{maskey2022generalized,zhang2021eigen}, which is strictly more expressive than the 1-Weisfeiler Leman (WL) test~\cite{morris2019weisfeiler,xu2018powerful}. To balance effectivness and efficiency, our main idea is to utilize random node features~\cite{ijcai2021p291,sato2021random}, which are known to enhance model expressiveness beyond 1-WL tests. Specifically, our encoder utilizes an MPNN $h_{\mathrm{PSE}}(\cdot)$ integrated with random node features $\boldsymbol{X}\sim\mathcal{N}(\boldsymbol{0},\boldsymbol{I})$, where $\mathcal{N}(\boldsymbol{0},\boldsymbol{I})$ denotes the Gaussian distribution:
\begin{equation}
\boldsymbol{Z}_{\mathrm{PSE},\tilde{c}}=h_{\mathrm{PSE}}(G _{\tilde{c}}^{\prime}),\boldsymbol{Z}_{\mathrm{PSE},\tilde{s}}=h_{\mathrm{PSE}}(G_{\tilde{s}}^{\prime}),
\label{eq:positionalstructuralinformation}
\end{equation}
where $G_{\tilde{c}}^\prime=(\boldsymbol{X},\mathcal{E}_{\tilde{c}})$ and $G_{\tilde{s}}^\prime=(\boldsymbol{X},\mathcal{E}_{\tilde{s}})$ represents the invariant and variant subgraphs with random node features, respectively. The edge sets $\mathcal{E}_{\tilde{c}}$ and $\mathcal{E}_{\tilde{s}}$ are kept the same as
$\boldsymbol{A}_{\tilde{c}}$ and $\boldsymbol{A}_{\tilde{s}}$, respectively. To guide the encoder maximally learn positional and structural information, we also design a predictor $w_{\mathrm{PSE}}(\cdot)$ to recover the hand-crafted position encodings as follows:
\begin{equation}
\mathcal{L}_{\mathrm{PSE}}=\mathbb{E}_{G_{\tilde{c}}}[\lvert z_{\mathrm{PSE}}-w_{\mathrm{PSE}}(\boldsymbol{Z}_{\mathrm{PSE},\tilde{c}})\rvert]+\mathbb{E}_{G_{\tilde{s}}}[\lvert z_{\mathrm{PSE}}-w_{\mathrm{PSE}}(\boldsymbol{Z}_{\mathrm{PSE},\tilde{s}})\rvert],
\end{equation}
where $\lvert\cdot\rvert$ denotes the $\ell_1$ loss and $\boldsymbol{z}_{\mathrm{PSE}}$ is the hand-crafted position encoding of the original unmasked graph.
This loss encourages the encoder to utilize the positional and structural information from the partially subgraph to approximate the PSE of the original graph. Note that $\boldsymbol{z}_{\mathrm{PSE}}$ can contain information from invariant subgraphs and lead to shortcuts. To prevent the introduction of spurious correlations into the disentangler, we have severed the back-propagation from the encoder to the disentangler.

With the learned node representation and PSEs of the invariant and variant subgraphs, we learn overall graph representations and predict graph labels as follows: 
\begin{equation}
\begin{aligned}
\boldsymbol{Z}^\prime_{\tilde{c}},\boldsymbol{Z}^\prime_{\tilde{s}}=&\mathrm{GT}^\prime([\boldsymbol{Z}_{\tilde{c}},\boldsymbol{Z}_{\mathrm{PSE},\tilde{c}}],[\boldsymbol{Z}_{\tilde{s}},\boldsymbol{Z}_{\mathrm{PSE},\tilde{s}}]),\\
\hat{y}_{\tilde{c}}=&w_{\tilde{c}}(\mathrm{Pooling}(\boldsymbol{Z}^\prime_{\tilde{c}})),\\
\hat{y}_{\tilde{s}}=&w_{\tilde{s}}(\mathrm{Pooling}(\boldsymbol{Z}^\prime_{\tilde{s}})),
\end{aligned}
\label{eq:graphrepresentation}
\end{equation}
where $\mathrm{GT}^\prime(\cdot)$ implements hybrid MPNN and Transformer layers to produce the overall invariant subgraph representation $\boldsymbol{Z}^\prime_{\tilde{c}}$ and variant subgraph representation $\boldsymbol{Z}^\prime_{\tilde{s}}$, similar to the approach in~\cite{wang2021causal}. $\mathrm{Pooling(\cdot)}$ is a global pooling function, $w_{\tilde{c}}(\cdot)$ and $w_{\tilde{s}}(\cdot)$ are two classifiers. Following~\cite{wudiscovering}, during training, we formulate the overall prediction as 
\begin{equation}\label{eq:interventionprediction}
    \hat{y}=\hat{y}_{\tilde{c}}\odot\sigma(\hat{y}_{\tilde{s}}),
\end{equation}
which mimics the intervention $do(S=\tilde{s})$ in causal theory. During the inference phase, we use $\hat{y}_{\tilde{c}}$ as the final prediction, which shields the influence of variant subgraph $G_{\tilde{s}}$.

\subsection{Invariant GT Learning Module}
In this section, we introduce our learning objectives.
Building on the GNN-based graph invariant learning literature ~\cite{wudiscovering,li2022learning}, we introduce interventional distributions to separate invariant and variant subgraphs and encourage OOD generalization. The loss function for graph invariant learning based on interventional distributions is as follows:
\begin{equation}
\begin{aligned}
\mathcal{L}_I=&\mathbb{E}_{G_s}[\mathcal{L}(h(\rv{G}),\rv{Y}|do(\rv{G}_S=G_s))] \\
&+ \lambda\mathrm{Var}_{G_s}(\{\mathcal{L}(h(\rv{G}),\rv{Y}|do(\rv{G}_S=G_s))\}),
\end{aligned}
\label{eq:intervention}
\end{equation}
where $\mathcal{L}(h(G),Y|do(\rv{G}_S=G_s))$ computes the loss function under the $s$-interventional distribution, $\mathrm{Var(\cdot)}$ calculates the variance of loss over different $s$-interventional distributions, and $\lambda$ is a hyper-parameter. As shown in causal learning literature~ \cite{pearl2009causality}, optimizing the intervention-based loss $\mathcal{L}_I$ can effectively encourage the independence between invariant and variant factors, and thus optimizes the objective function in Eq.~\eqref{eq:GTObj}.

Besides, we also have a normal classification loss function for the classifiers $w_{\tilde{s}}(\cdot)$ as:
\begin{equation}
    \mathcal{L}_{\tilde{S}}=\mathbb{E}_{\tilde{s}}[\ell(\hat{y}_{\tilde{s}},y)],
\end{equation}
where $\ell(\cdot)$ denotes the loss function on a single instance. Incorporating the entropy loss $\mathcal{L}_E$ and the encoding loss $\mathcal{L}_{\mathrm{PSE}}$, the final optimization objective is defined as follows:
\begin{equation}
\min_{h,w_{\tilde{s}},w_{\mathrm{PSE}}} \mathcal{L}_I+ \alpha_S\mathcal{L}_{\tilde{S}}+\alpha_E\mathcal{L}_E+\alpha_{\mathrm{PSE}}\mathcal{L}_{\mathrm{PSE}},
\label{eq:finalobjective}
\end{equation}
where $\alpha_S,\alpha_E,\alpha_{\mathrm{PSE}}$ are hyper-parameters to control the impacts of different components. During the inference phase, we first calculate $\boldsymbol{E}_G$, and then use it to calculate $t_G$ by Eq.~\eqref{eq:testtimetraining}, and finally calculate $\hat{y}_{\tilde{c}}$ as the  prediction.

\subsection{Discussions}\label{subsec:discussion}

\begin{algorithm}[t]
  \SetAlgoLined
  \SetKwInOut{Input}{Input}\SetKwInOut{Output}{Output}
  \Input{A graph dataset $\graph=\{(G_i,Y_i)\}^{N}_{i=1}$}
  \Output{An optimized predictor $h(\cdot):\rv{G}\rightarrow \rv{Y}$}
  Calculate hand-crafted PSEs $\mathcal{Z}=\{z_{\mathrm{PSE},i}\}^{N}_{i=1}$ for $\graph$\;
  \For{jointly sampled minibatch $\mathcal{B}$ from $\mathcal{G}$ and $\mathcal{Z}$}{
    \For{each graph instance $(G,Y,z_{\mathrm{PSE}})\ \in \mathcal{B}$}{
        Calculate the attention $\boldsymbol{Z}_{\mathrm{Attn},{\tilde{c}}}$ and $\boldsymbol{Z}_{\mathrm{Attn},{\tilde{s}}}$ by Eq.~\eqref{eq:complementatt}  \;
        Obtain the adjacency matrix $\boldsymbol{A}_{\tilde{c}}$ and $\boldsymbol{A}_{\tilde{s}}$ by Eq.~\eqref{eq:subgraphadjacencymatrix}  \;
        Compute the subgraph node representations $\boldsymbol{Z}_{\mathrm{MPNN},{\tilde{c}}}$ and $\boldsymbol{Z}_{\mathrm{MPNN},{\tilde{s}}}$ by Eq.~\eqref{eq:mpnnoutput}  \;
        Aggregate the node representations to obtain $\boldsymbol{Z}_{\tilde{c}}$ and $\boldsymbol{Z}_{\tilde{s}}$ by Eq.~\eqref{eq:mlpaggregate}  \;
        Calculate learned PSEs $\boldsymbol{Z}_{\mathrm{PSE},\tilde{c}}$ and $\boldsymbol{Z}_{\mathrm{PSE},\tilde{s}}$ by Eq.~\eqref{eq:positionalstructuralinformation} \;
        Predict subgraph labels $\hat{y}_{\tilde{c}}$ and $\hat{y}_{\tilde{s}}$ by Eq.~\eqref{eq:graphrepresentation} and the overall prediction by Eq.~\eqref{eq:interventionprediction}\;
        }
    Calculate the objective function by Eq.~\eqref{eq:finalobjective} \;
    Update model parameters using back propagation \;
    }
  \caption{The training procedure of \name}
  \label{algorithm: pseudocode}
\end{algorithm}

We present the pseudo-code of \name~in Algorithm~\ref{algorithm: pseudocode}. Additionally, the time complexity of our \name~is $O(\lvert\mathcal{E}\rvert d+\lvert\mathcal{V}\rvert^2d^2)$, where $\lvert \mathcal{V}\rvert$ and $\lvert \mathcal{E}\rvert$ denote the number of nodes and edges, respectively, and $d$ is the dimensionality of the representations. Specifically, we employ a combination of MPNN and Transformer layers to implement \name, resulting in a complexity of $O(\lvert\mathcal{E}\rvert d+\lvert\mathcal{V}\rvert^2d^2)$. The time complexity during test-time training, which involves iterative computation of the softmax function, is $O(\lvert\mathcal{V}\rvert^2T)$, where $T$ denotes the number of iterations. Since $T$ is a small constant, the overall time complexity remains $O(\lvert\mathcal{E}\rvert d+\lvert\mathcal{V}\rvert^2d^2)$. In comparison,  the complexity of MPNNs is $O(\lvert\mathcal{E}\rvert d+\lvert\mathcal{V}\rvert d^2)$ and the time complexity of representative GTs is $O(\lvert\mathcal{V}\rvert^2d^2)$. Therefore, the complexity of \name~is on the same scale as that of the existing GT methods. We also evaluate the computational overhead introduced by \name's architecture and auxiliary objectives in Appendix~\ref{subsec:runningtime}.

%% file: sections/4-Exp.tex
\section{Experiments}\label{sec:exp}
In this section, we conduct extensive experiments to evaluate our propose \name~on both synthetic and real-world datasets, with the aim of answering the following research questions: 
\begin{itemize}
    \item[\textbf{RQ1}:] Can \name~effectively improve the OOD generalizations of GTs?
    \item[\textbf{RQ2}:] Can our \name~identify invariant subgraphs?
    \item[\textbf{RQ3}:] How do the key components contribute to the results?
    \item[\textbf{RQ4}:] Is \name~sensitive across various hyper-parameter settings?
\end{itemize}

\subsection{Experimental Setup}

\subsubsection{Datasets}
We employ a synthetic dataset (GOOD-Motif) and four real-world datasets (GOOD-HIV, DrugOOD, GOOD-SST2, and GOOD-Twitter) for the graph classification task~\cite{gui2022good}. For the synthetic dataset, we use both the base and size splits of GOOD-Motif. For molecular property prediction tasks, we apply the scaffold and size splits of GOOD-HIV, and the assay split of DrugOOD to evaluate the performance of our method across various shifts. Additionally, GOOD-SST2 and GOOD-Twitter are graph-based natural language processing datasets, characterized by variations in sentence lengths. We report accuracy for
GOOD-Motif, GOOD-SST2 and GOOD-Twitter, and ROC-AUC scores for GOOD-HIV and DrugOOD. The dataset statistics are presented in Table~\ref{tab:datasetstatistic}. More details on the datasets are in Appendix~\ref{sec:dataset-detail}.

\begin{table}
\centering
    \caption{Statistics of datasets.}
    \label{tab:datasetstatistic}
    \begin{tabular}{lccc} \toprule 
         \multirow{2}{*}{Dataset}   & \multicolumn{3}{c}{Number of Graphs} \\ 
         &Train& Validation& Test\\  \midrule  
         GOOD-Motif-basis/size&18,000&3,000&3,000\\
         GOOD-HIV-scaffold&24,682&4,113&4,108\\
         GOOD-HIV-size&26,169&4,112&3,961\\
         GOOD-SST2-length&24,744&17,206&17,490\\
         GOOD-Twitter-length&2,590&1,785&1,457\\
         DrugOOD-assay&34,179&19,028&19,032\\\bottomrule
    \end{tabular}
\end{table}

\subsubsection{Compared Methods}
We compare our \name~with representative state-of-the-art methods from three groups. The first group is general invariant learning methods, including standard ERM, IRM~\cite{arjovsky2019invariant}, and VREx~\cite{krueger2021out}. The second group is graph OOD methods including DIR~\cite{wudiscovering}, GREA~\cite{liu2022graph}, GIL~\cite{li2022learning}, GSAT~\cite{miao2022interpretable}, CIGA~\cite{chen2022learning}, iMoLD~\cite{zhuang2023learning}, and CAL+~\cite{sui2024enhancing}. The third group is graph Transformer methods including Graphormer~\cite{ying2021transformers}, GraphGPS~\cite{rampavsek2022recipe}, and EXPHORMER~\cite{shirzad2023exphormer}. More details on the baselines are in Appendix~\ref{sec:baseline-detail}.

\subsubsection{Optimization and Hyper-parameters}
 For $\mathrm{GT}^M(\cdot)$ employing the GraphGPS layer \cite{rampavsek2022recipe}, along with MPNN $h_{\mathrm{PSE}}(\cdot)$ and $\mathrm{GT}^\prime(\cdot)$, the numer of the respective layers is 2, 3 and 1, respectively. For all GTs, the dimensionality of both graph-level and node-level representations is $d=128$, whereas for GNN-based baselines, $d=300$. The global pooling function employed is mean pooling. The hand-crafted PSE $z_{\mathrm{PSE}}$ is Laplacian eigenvectors encoding (LapPE). 
The hyper-parameter $\lambda$ in Eq.~\eqref{eq:intervention} is chosen from $\{10^2,10^1,10^0,10^{-1}\}$. The hyper-parameter $\alpha_E$ in Eq.~\eqref{eq:finalobjective} is chosen from $\{10^{-1},10^{-2}\}$. The hyper-parameter $\alpha_{\mathrm{PSE}}$ and $\alpha_S$ in Eq.~\eqref{eq:finalobjective} are set at $10^{-2}$ and $10^0$, repectively. We report the mean results and standard deviations of three runs. For the number of epochs, we use 100 for our method and the GT baselines. An exception is the GOOD-Motif basis split, which is limited to 50 epochs to prevent overfitting. Other GNN-based baselines adhere to the GOOD benchmark~\cite{gui2022good} and their respective original papers. We adopt Adaptive Moment Estimation (Adam) for optimization. The activation function is ReLU~\cite{agarap2018deep}. The selected hyper-parameters are reported in Table \ref{tab:hyper-parameter}. Besides, our hyper-parameters have concrete meanings and do not necessitate a resource-intensive tuning:
\begin{itemize}[leftmargin=1em]
    \item Regarding $\alpha_S$: This parameter solely controls the influence of the classification loss for the classifier $w_{\tilde{s}}(\cdot)$. Actually, in our experiments, $\alpha_S$ is consistently set to 1.
    \item Regarding $\alpha_E,\alpha_{\mathrm{PSE}}$: They can be relatively straightforwardly determined based on the relative magnitudes of each loss component. Specifically, we can let $\alpha_E\mathcal{L}_E$ and $\alpha_{\mathrm{PSE}}\mathcal{L}_{\mathrm{PSE}}$ to be of comparable scale, while ensuring $\mathcal{L}_I$ dominates both.
\end{itemize}

\begin{table*}
\centering
    \caption{The chosen hyper-parameters of $\lambda$, $\alpha_E$ and $\alpha_{\mathrm{PSE}}$ on each dataset. }
    \label{tab:hyper-parameter}
    \begin{tabular}{lccccccc} \toprule 
         &\multicolumn{2}{c}{GOOD-Motif}&\multicolumn{2}{c}{GOOD-HIV}&GOOD-SST2&GOOD-Twitter&DrugOOD\\ \cline{2-8} 
         &basis&size&scaffold&size&length&length&assay\\ \midrule 
         $\lambda$       &10&1&0.1&100&0.1&100&100\\ 
         $\alpha_E$      &0.1&0.1&0.1&0.01&0.01&0.01&0.01\\  
         $\alpha_{\mathrm{PSE}}$      &0.01&0.01&0.01&0.01&0.01&0.01& 0.01\\ \bottomrule
    \end{tabular}
\end{table*}

\subsection{Performance Comparisons}\label{subsec:performance}
To answer \textbf{RQ1}, we evaluate the performance of \name~on all datasets. The metrics for ERM, IRM, VREx, DIR, GSAT and CIGA on the GOOD-Motif, GOOD-HIV, and GOOD-SST2 datasets are obtained from the original benchmark~\cite{gui2022good}. The results are shown in Table \ref{tab: graph classification results}. Our observations are as follows.

The \name ~ model consistently and significantly outperforms the baselines, achieving superior performance across all datasets. These results indicate that our proposed method effectively manages graph distribution shifts and exhibits an exceptional out-of-distribution generalization ability.

\begin{table*}
    \caption{The graph classification results (\%) on the testing sets. We report accuracy for GOOD-Motif, GOOD-SST2 and GOOD-Twitter, and ROC-AUC scores for GOOD-HIV and DrugOOD. Numbers after ± indicate variances. The best results are in \best{bold} and the second-best results are \secbest{underlined}. }
    \centering
    \label{tab: graph classification results}
    \resizebox{\textwidth}{!}{
    \begin{tabular}{lccccccc} \toprule 
         &\multicolumn{2}{c}{GOOD-Motif}&\multicolumn{2}{c}{GOOD-HIV}&GOOD-SST2&GOOD-Twitter&DrugOOD\\ \cline{2-8} 
         &basis&size&scaffold&size&length&length&assay\\ \midrule 
         GNN+ERM       &63.80\scalebox{0.75}{±10.36}& 53.46\scalebox{0.75}{±4.08}&69.55\scalebox{0.75}{±2.39}&59.19\scalebox{0.75}{±2.29}&80.52\scalebox{0.75}{±1.13}&55.14\scalebox{0.75}{±1.77}& 71.48\scalebox{0.75}{±1.01}\\ 
         GNN+IRM       &59.93\scalebox{0.75}{±11.46}& 53.68\scalebox{0.75}{±4.11}&\secbest{70.17}\scalebox{0.75}{±2.78}&59.94\scalebox{0.75}{±1.59}&80.75\scalebox{0.75}{±1.17}&55.46\scalebox{0.75}{±0.69}& 71.63\scalebox{0.75}{±1.15}\\  
         GNN+VREx      &66.53\scalebox{0.75}{±4.04} & 54.47\scalebox{0.75}{±3.42}&69.34\scalebox{0.75}{±3.54}&58.49\scalebox{0.75}{±2.28}&80.20\scalebox{0.75}{±1.39}&56.87\scalebox{0.75}{±0.96}& 71.72\scalebox{0.75}{±0.92}\\ \midrule  
         DIR       &39.99\scalebox{0.75}{±5.50} & 44.83\scalebox{0.75}{±4.00}&68.44\scalebox{0.75}{±2.51}&57.67\scalebox{0.75}{±3.75}&81.55\scalebox{0.75}{±1.06}&55.91\scalebox{0.75}{±2.20}& 70.28\scalebox{0.75}{±0.76}\\
         GREA       &52.49\scalebox{0.75}{±5.94} & 57.74\scalebox{0.75}{±0.05}&69.05\scalebox{0.75}{±2.36}&\secbest{62.59}\scalebox{0.75}{±1.92}&81.81\scalebox{0.75}{±0.29}&57.15\scalebox{0.75}{±1.76}& 71.68\scalebox{0.75}{±1.07}\\
         GIL       &54.22\scalebox{0.75}{±9.83} & 48.68\scalebox{0.75}{±8.56}&67.18\scalebox{0.75}{±3.67}&62.00\scalebox{0.75}{±0.81}&82.19\scalebox{0.75}{±1.10}&58.50\scalebox{0.75}{±0.49}& 71.13\scalebox{0.75}{±0.68}\\
         GSAT      &55.13\scalebox{0.75}{±5.41} & 60.76\scalebox{0.75}{±5.94}&70.07\scalebox{0.75}{±1.76}&60.73\scalebox{0.75}{±2.39}&81.49\scalebox{0.75}{±0.76}&54.89\scalebox{0.75}{±1.43}&\secbest{71.75}\scalebox{0.75}{±0.54}\\  
         CIGA      &67.15\scalebox{0.75}{±8.19} & 54.42\scalebox{0.75}{±3.11}&69.40\scalebox{0.75}{±1.97}&59.55\scalebox{0.75}{±2.56}&80.46\scalebox{0.75}{±2.00}&56.14\scalebox{0.75}{±0.90}& 70.51\scalebox{0.75}{±0.83}\\
         iMoLD     &\secbest{71.02}\scalebox{0.75}{±1.90} & 51.81\scalebox{0.75}{±6.45}&69.06\scalebox{0.75}{±2.36}&62.20\scalebox{0.75}{±2.59}&79.41\scalebox{0.75}{±0.65}&57.08\scalebox{0.75}{±2.21}& 71.66\scalebox{0.75}{±0.05}\\
         CAL+     &63.86\scalebox{0.75}{±3.14} & \secbest{72.22}\scalebox{0.75}{±4.78}&70.03\scalebox{0.75}{±1.73}&61.17\scalebox{0.75}{±5.54}&80.78\scalebox{0.75}{±0.07}&58.78\scalebox{0.75}{±1.08}& 71.64\scalebox{0.75}{±1.54}\\\midrule 
         Graphormer&40.21\scalebox{0.75}{±1.55} & 53.22\scalebox{0.75}{±3.83}&69.38\scalebox{0.75}{±2.64}&62.26\scalebox{0.75}{±1.22}&82.10\scalebox{0.75}{±0.70}&58.25\scalebox{0.75}{±1.41}& 69.48\scalebox{0.75}{±0.57}\\  
         GraphGPS  &46.48\scalebox{0.75}{±1.80} & 59.90\scalebox{0.75}{±5.22}&67.90\scalebox{0.75}{±3.87}&59.23\scalebox{0.75}{±3.19}&81.91\scalebox{0.75}{±0.80}&\secbest{58.89}\scalebox{0.75}{±1.32}& 71.57\scalebox{0.75}{±0.86}\\  
         EXPHORMER &46.44\scalebox{0.75}{±11.81}& 65.62\scalebox{0.75}{±3.27}&69.24\scalebox{0.75}{±2.22}&53.85\scalebox{0.75}{±2.61}&\secbest{82.76}\scalebox{0.75}{±0.66}&58.71\scalebox{0.75}{±1.59}& 70.88\scalebox{0.75}{±0.42}\\ \midrule 
         \name     &\best{81.97}\scalebox{0.75}{±16.74}& \best{80.02}\scalebox{0.75}{±7.94}&\best{70.38}\scalebox{0.75}{±0.27}&\best{62.69}\scalebox{0.75}{±0.11}&\best{83.07}\scalebox{0.75}{±0.93}&\best{59.42}\scalebox{0.75}{±0.51}& \best{71.75}\scalebox{0.75}{±0.08}\\ \bottomrule
    \end{tabular}
    }
\end{table*}

The synthetic dataset GOOD-Motif serves as a crucial benchmark for evaluating structural shift resilience. On the basis split, \name~improves classification accuracy by 11.0\% over the strongest baselines, and by 7.8\% on the size split. Furthermore, while other graph Transformer methods surpass most general invariant learning methods and graph OOD methods on the size split due to architectural advantages, their OOD generalization ability on both the basis and size splits is significantly inferior compared to our \name. Our \name~achieves the best performance on the synthetic dataset, demonstrating its superior capability to effectively handle structural shifts.

Besides, our method achieves the best performance across all four real-world datasets. This performance underscores \name's strong ability to effectively manage distribution shifts in real-world graph scenarios. Besides, different datasets have different distribution shifts, e.g., GOOD-HIV is split based on scaffold and size, DrugOOD has a distribution shift related to assay types, and features of GOOD-SST2 have varying sentence lengths. Consequently, the results demonstrate that our proposed method effectively manages diverse types of distribution shifts in real-world graph datasets.

\begin{figure*}
    \centering
        \begin{subfigure}{0.19\linewidth}
        \centering
        \includegraphics[width=\linewidth]{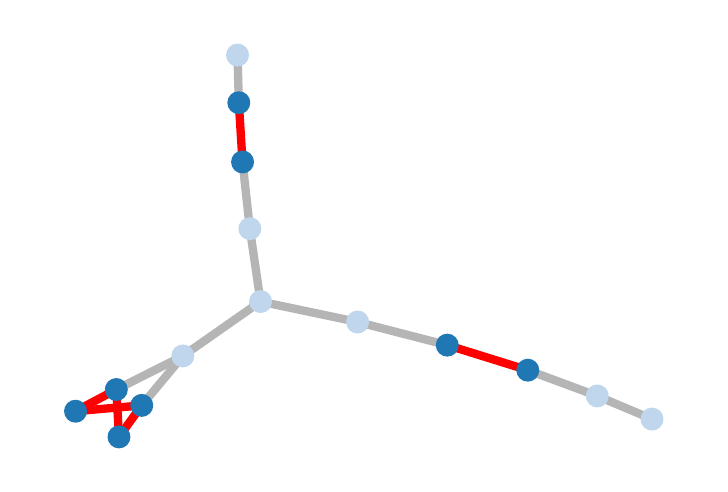}
        \caption{DIR}
        \end{subfigure}
        \begin{subfigure}{0.19\linewidth}
        \centering
        \includegraphics[width=\linewidth]{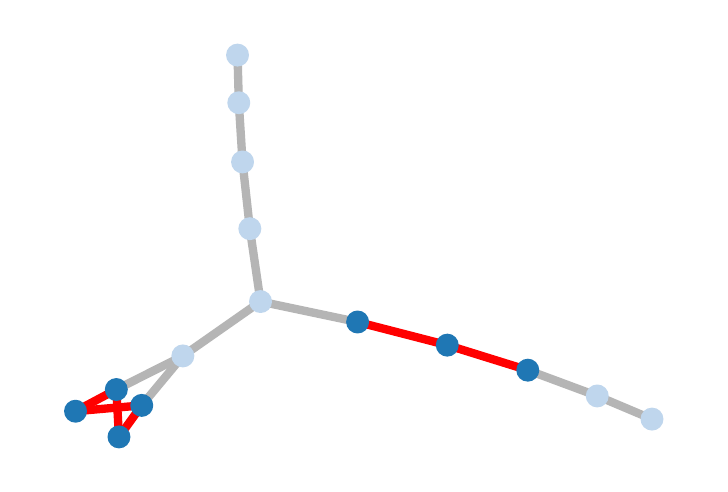}
        \caption{GSAT}
        \end{subfigure}
        \begin{subfigure}{0.19\linewidth}
        \centering
        \includegraphics[width=\linewidth]{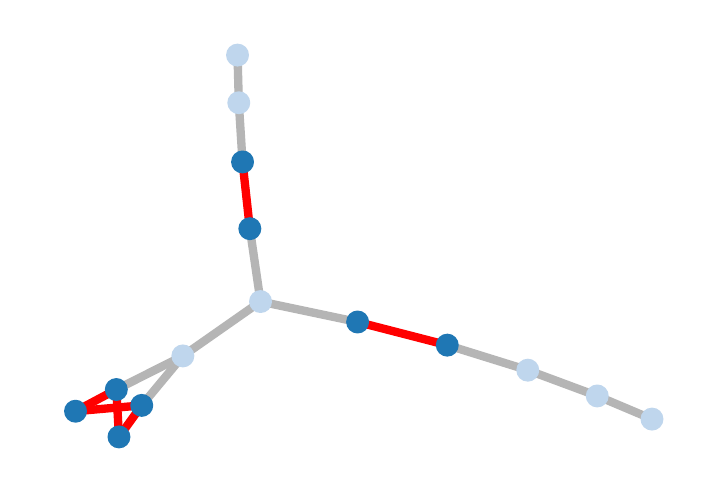}
        \caption{CIGA}
        \end{subfigure}
        \begin{subfigure}{0.19\linewidth}
        \centering
        \includegraphics[width=\linewidth]{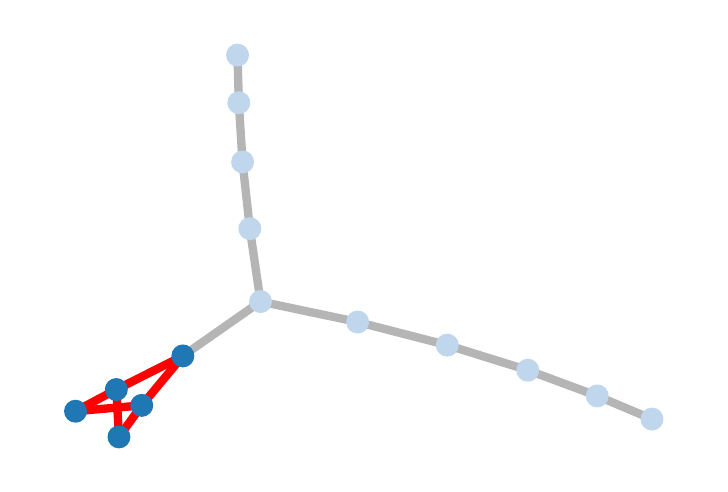}
        \caption{\name}
        \end{subfigure}
        \begin{subfigure}{0.19\linewidth}
        \centering
        \includegraphics[width=\linewidth]{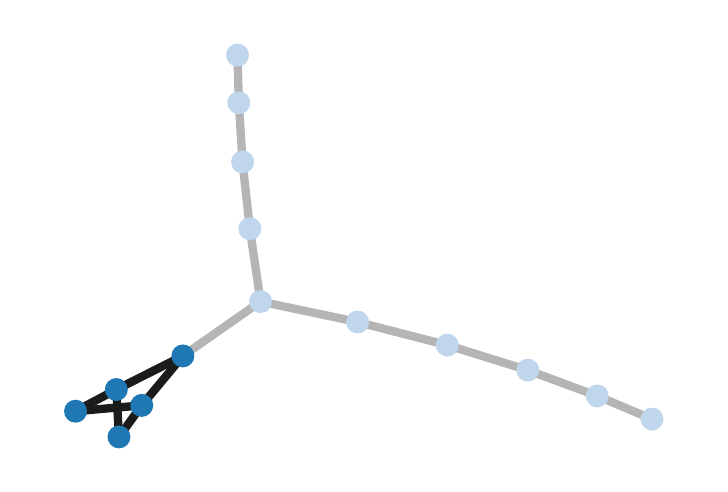}
        \caption{Ground Truth}
        \end{subfigure}
        \setcounter{subfigure}{0}
    \centering
        \begin{subfigure}{0.19\linewidth}
        \centering
        \includegraphics[width=\linewidth]{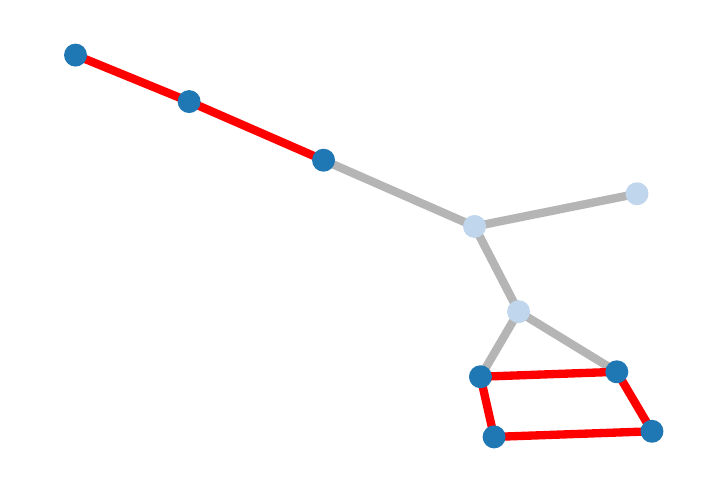}
        \caption{DIR}
        \end{subfigure}
        \begin{subfigure}{0.19\linewidth}
        \centering
        \includegraphics[width=\linewidth]{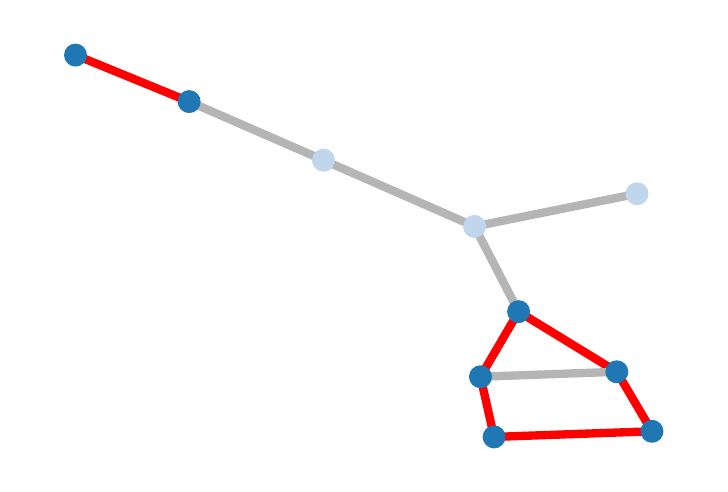}
        \caption{GSAT}
        \end{subfigure}
        \begin{subfigure}{0.19\linewidth}
        \centering
        \includegraphics[width=\linewidth]{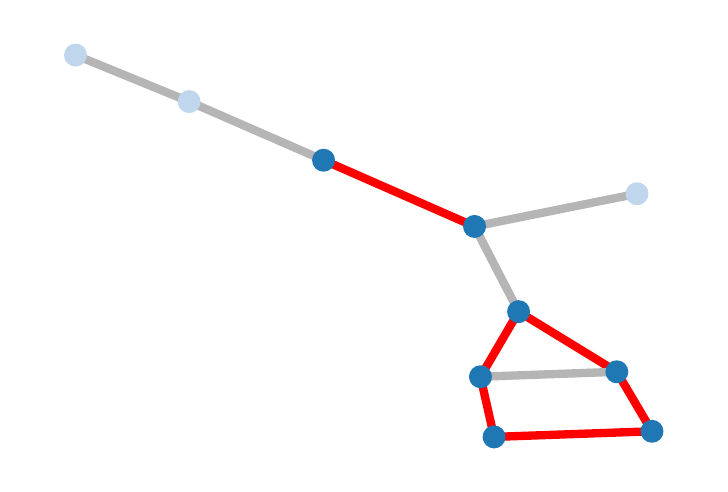}
        \caption{CIGA}
        \end{subfigure}
        \begin{subfigure}{0.19\linewidth}
        \centering
        \includegraphics[width=\linewidth]{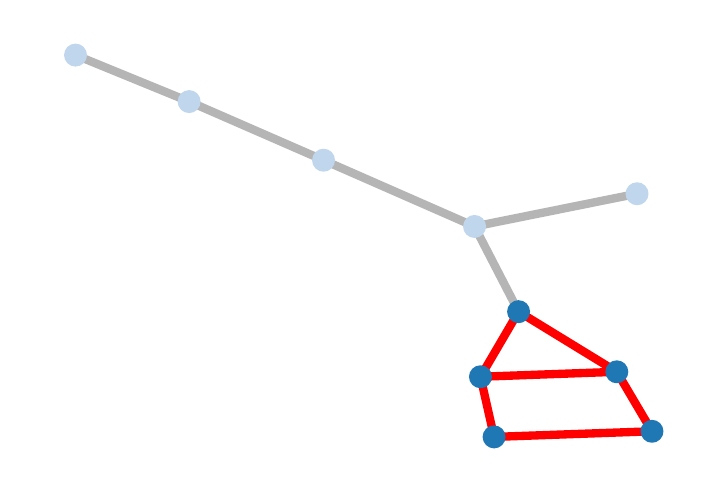}
        \caption{\name}
        \end{subfigure}
        \begin{subfigure}{0.19\linewidth}
        \centering
        \includegraphics[width=\linewidth]{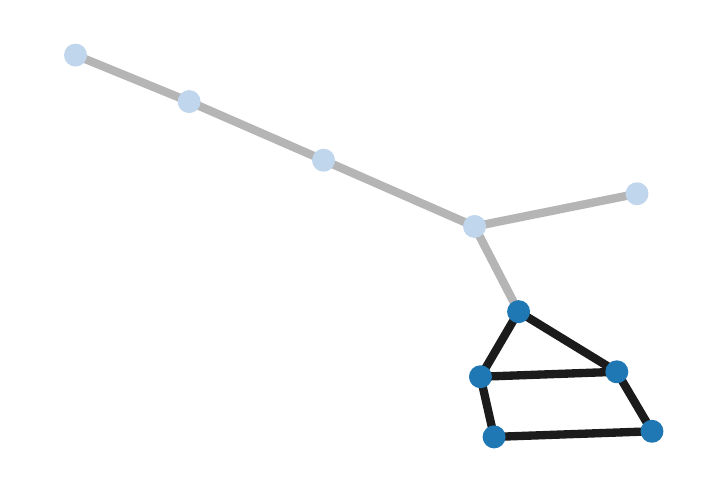}
        \caption{Ground Truth}
        \end{subfigure}
    \caption{Visualizations of the learned invariant subgraphs from the testing set of the GOOD-Motif dataset with basis split. In Figures (a) to (d), the invariant subgraphs identified by different methods are highlighted in red, while the ground truth is depicted in black in Figure (e).}
    \label{fig:visualization}
\end{figure*}

To answer \textbf{RQ2}, we further conduct a comparative study to evaluate whether our proposed method can accurately identify invariant subgraphs. Specifically, we compare \name~with baseline methods that also explicitly generate subgraphs via edge masking, using the ground-truth invariant subgraphs as a reference on the basis split of the GOOD-Motif dataset. Both qualitative and quantitative evaluations are performed, with Precision@10 adopted as the evaluation metric. The results, summarized in Table~\ref{tab:invariantsubgraphdiscovering} and visualized in Figure~\ref{fig:visualization}, demonstrate that \name~significantly outperforms the baselines in capturing invariant subgraphs, laying the foundation for its efficacy in handling distribution shifts.

\begin{table}
    \caption{The Precision@10 of discovering the ground-truth invariant subgraphs on the basis split of GOOD-Motif.}
    \label{tab:invariantsubgraphdiscovering}
    \begin{tabular}{lc} \toprule 
         Methods&Precision@10\\  \midrule  
         DIR&0.203\scalebox{0.75}{±0.041}\\
         GSAT&0.396\scalebox{0.75}{±0.077}\\
         CIGA&0.247\scalebox{0.75}{±0.024}\\
         \name&\best{0.698}\scalebox{0.75}{±0.069}\\\bottomrule
    \end{tabular}
\end{table}

\subsection{Ablation Studies}\label{subsec:ablation}
\begin{table}
    \caption{Ablation studies on \name. The "w/o entropy" row indicates the removal of entropy loss and test-time training. The "+GIL" row represents the substitution of the disentangler with GIL's subgraph generator. The "w/o encoder" row represents the complete removal of the encoder. The "+LapPE" and "+SignNet" rows indicate Laplacian PE and SignNet encodings of the input graphs.
    }
    \resizebox{\linewidth}{!}{
    \begin{tabular}{lcccc} \toprule 
         &\multicolumn{2}{c}{GOOD-Motif}&\multicolumn{2}{c}{GOOD-HIV}\\ \cline{2-5} 
         &basis&size&scaffold&size\\ \midrule  
         \name&\best{81.97}\scalebox{0.75}{±16.74}& \best{80.02}\scalebox{0.75}{±7.94}&\best{70.38}\scalebox{0.75}{±0.27}&\best{62.69}\scalebox{0.75}{±0.11}\\ 
         w/o entropy&40.40\scalebox{0.75}{±3.22}&69.84\scalebox{0.75}{±6.48}&62.67\scalebox{0.75}{±4.93}&56.71\scalebox{0.75}{±4.00}\\ 
         +GIL&48.08\scalebox{0.75}{±4.20}&79.63\scalebox{0.75}{±7.08}&68.19\scalebox{0.75}{±1.92}&61.30\scalebox{0.75}{±0.69}\\
         w/o encoder&52.32\scalebox{0.75}{±12.56}&71.59\scalebox{0.75}{±5.13}&65.79\scalebox{0.75}{±3.01}&59.18\scalebox{0.75}{±2.71} \\ 
         +LapPE&59.29\scalebox{0.75}{±3.57}&67.80\scalebox{0.75}{±4.67}&67.11\scalebox{0.75}{±7.05}&58.45\scalebox{0.75}{±1.10} \\ 
         +SignNet&63.83\scalebox{0.75}{±16.77}&48.51\scalebox{0.75}{±9.65}&63.07\scalebox{0.75}{±1.75}&56.71\scalebox{0.75}{±3.79} \\ 
        \bottomrule
    \end{tabular}
    }
    \label{tab: ablation study}
\end{table}

To answer \textbf{RQ3}, we evaluate two key modules of \name: \disentangler~and \encoder. For the \disentangler, we remove the entropy loss and test-time training, retaining only the standard attention for disentanglement (denoted as w/o entropy), and then replace it with the subgraph generator from GIL (denoted as +GIL). For the \encoder, we remove it entirely (denoted w/o encoder) and then introduce two hand-crafted encodings based on the origin graphs at the beginning of the model (LapPE and SignNet~\cite{lim2022sign}). The results based on GOOD-Motif and GOOD-HIV are shown in Table \ref{tab: ablation study}, while other datasets show similar patterns. We have the following observations:

The full model \name~always achieves the best scores compared with the five variants, indicating that each module is necessary for OOD generalization and jointly optimizing these modules leads to the desired result.

By removing the entropy loss and test-time training, our method achieves suboptimal performance. This suggests that the entropy loss and test-time training significantly contribute to the disentangler's ability to separate invariant and variant subgraphs, thereby enhancing the OOD generalization performance. Furthermore, substituting the disentangler with GIL‘s subgraph generator similarly yields suboptimal results, underscoring the architectural importance of the disentangler module.

Without the \encoder, our method also experiences a decline in performance, indicating that our specically designed encoder enhances the model's expressive capabilities while encouraging OOD generalization. Introducing the LapPE or SignNet based on the original graphs at the beginning of the model does slightly improve the results on the basis split of GOOD-Motif, but it falls short of the performance achieved with the full model. This underscores the importance of providing precise PSEs for subgraphs. Simply incorporating hand-crafted encodings of the original graphs can introduce spurious correlations between the graph structure and the labels, thereby reducing the model's OOD generalization capabilities.

\subsection{Hyper-parameter Sensitivity}
\begin{figure*}
\centering
	\begin{subfigure}{0.33\linewidth}
		\centering
		\includegraphics[width=\linewidth]{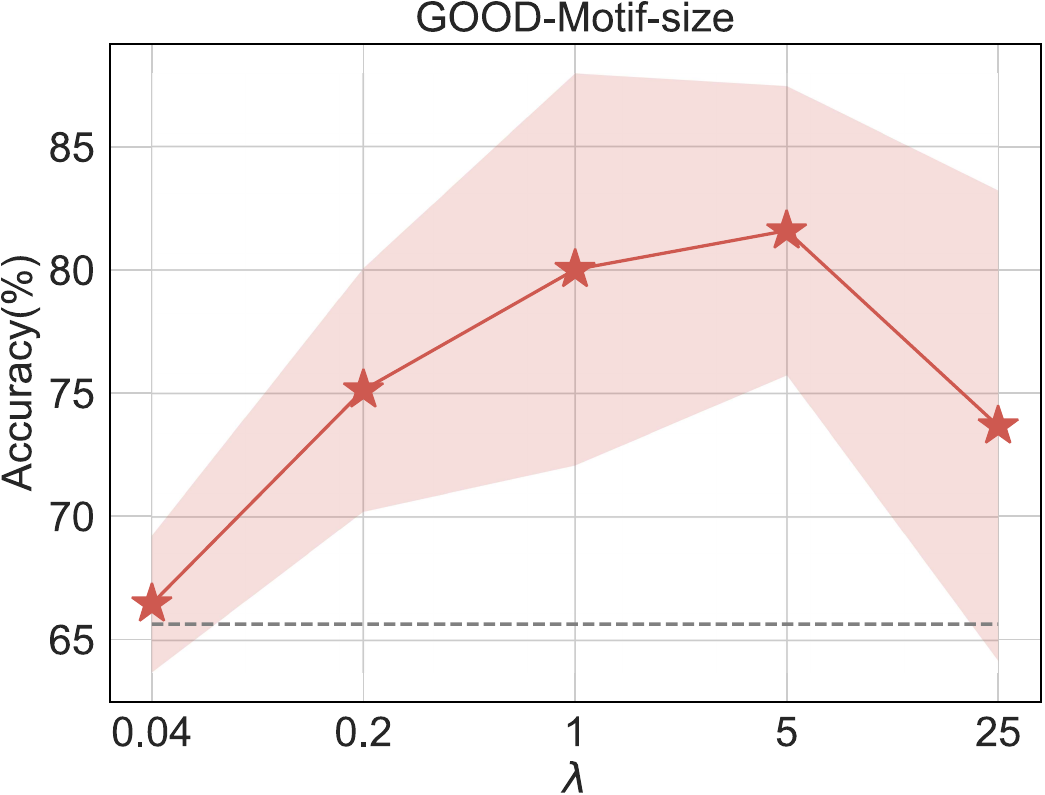}
	\end{subfigure}
	\begin{subfigure}{0.33\linewidth}
		\centering
		\includegraphics[width=\linewidth]{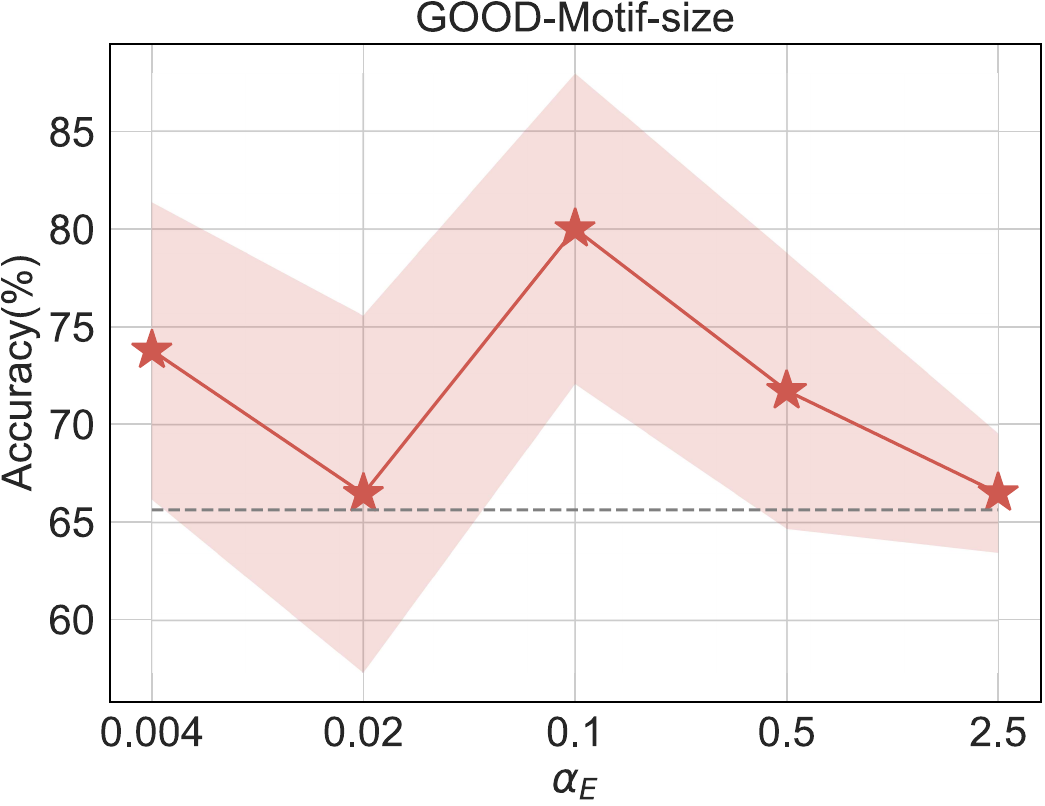}
	\end{subfigure}
	\begin{subfigure}{0.33\linewidth}
		\centering
		\includegraphics[width=\linewidth]{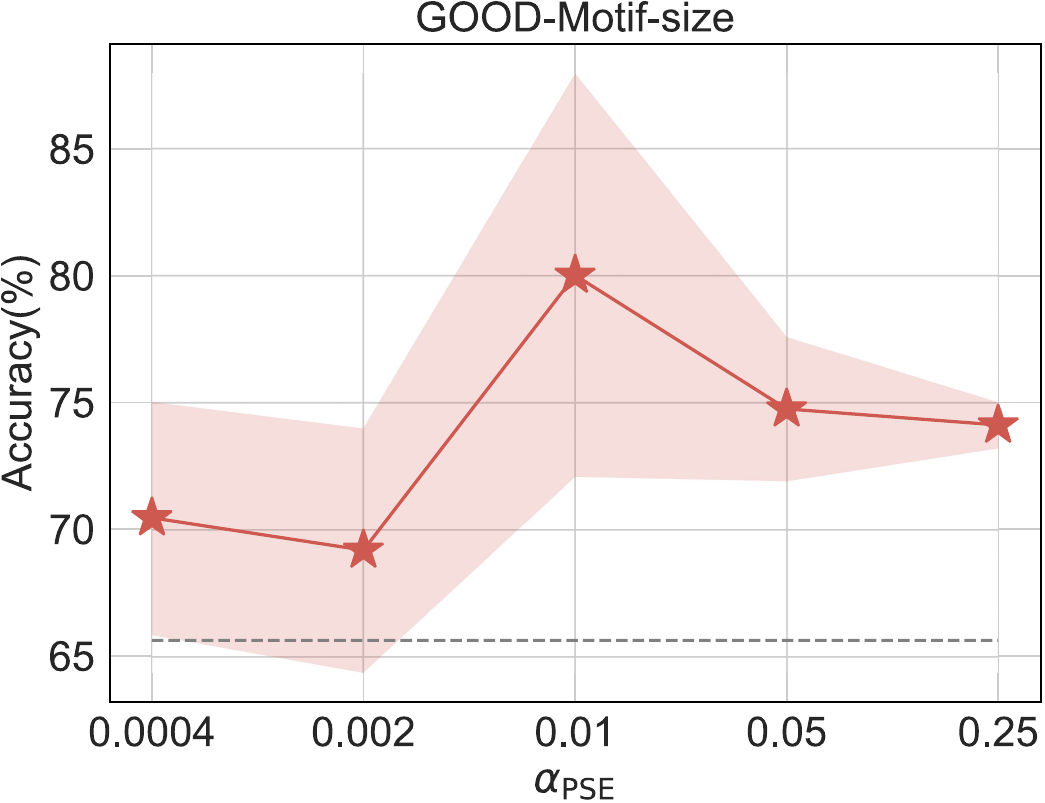}
	\end{subfigure}
    \caption{The impact of different hyper-parameters. Red lines denote the results of \name~and grey dashed lines correspond to the results of EXPHORMER.}
    \label{fig:hyper-parameter sensitivity}
\end{figure*}

To answer \textbf{RQ4}, we evaluate the sensitivity of hyper-parameters, including the invariant learning hyper-parameter $\lambda$, the entropy loss hyper-parameter $\alpha_E$ and the encoding loss hyper-parameter $\alpha_{\mathrm{PSE}}$. For simplicity, we only report the results on the size split of GOOD-Motif in Figure \ref{fig:hyper-parameter sensitivity}, while results on other datasets show similar patterns and are provided in Appendix~\ref{subsec:hyper-parameter}.

The invariant learning hyper-parameter $\lambda$ is crucial. Setting $\lambda$ too high can lead to an excessive emphasis on invariance, causing other loss to be overshadowed. Conversely, a too small $\lambda$ may be insufficient to differentiate invariant and variant subgraphs. The entropy loss hyper-parameter $\alpha_E$ is also important. Setting $\alpha_E$ too high can cause the model to favor invariant parts with higher certainty, neglecting other components. Conversely, if $\alpha_E$ is too low, it fails to maintain a sharp softmax function. The hyper-parameter of the encoding loss $\alpha_{\mathrm{PSE}}$ moderately impacts model performance, suggesting the need for a careful balance between the encoding loss and other losses. Although an appropriate choice can further improve the performance, our method is not very sensitive to hyper-parameters and outperforms the best GT with a wide range of hyper-parameters choices.

%% file: sections/5-Related.tex
\section{Related Work}\label{sec:related}

\subsection{Graph Transformers}
Transformers have seen tremendous success in various fields, including natural language processing (NLP)~\cite{tunstall2022natural}, computer vision~\cite{han2022survey}, and more recently, in the graph domain~\cite{shehzad2024graph,min2022transformer}. Compared to existing message-passing neural networks (MPNNs)~\cite{gilmer2017neural,kipf2022semi,velivckovic2018graph}, they show promise in addressing challenges like over-smoothing and over-squashing~\cite{cai2023connection,xing2024less} as well as capturing long-term dependencies~\cite{wu2021representing}. Over time, various GTs approaches have emerged. Graphormer~\cite{ying2021transformers} was among the first, using a single dense attention mechanism while adding structural features through centrality and spatial encodings. SAN~\cite{kreuzer2021rethinking} employed dual attention mechanisms on both fully-connected graphs and original graph edges, with Laplacian Positional Encodings for nodes. Subsequently, GraphTrans~\cite{wu2021representing} marked the introduction of a hybrid architecture, initially stacking MPNN layers before fully connecting the graph. GraphGPS~\cite{rampavsek2022recipe} provided a flexible framework for integrating message-passing networks with attention mechanisms, allowing for diverse combinations of PSEs. The field continues to evolve with recent GT variants such as Exphormer~\cite{shirzad2023exphormer}, Polynormer~\cite{dengpolynormer}, NAGphormer~\cite{chennagphormer}, among others~\cite{wu2024simplifying,zhang2022hierarchical,liu2024gradformer}. However, all of these GTs rely on the assumption that the training and testing graph comes from the same distribution, leaving their effectiveness under distribution shifts unexplored. To our knowledge, we are the first to study distribution shifts of GTs.

\subsection{Graph Invariant Learning}
To tackle the distribution shift problem of graph data, graph invariant learning has received widespread attention, aiming to harness invariance in graph data for OOD generalization~\cite{li2022out}. These approaches generally consider distribution shifts between training and testing data as stemming from unknown environmental variables~\cite{yuan2024environment,liu2022graph,gui2024joint}, focusing on invariant aspects of the data for prediction~\cite{li2022out,wudiscovering}. GIL~\cite{li2022learning} captures invariant relationships between graph structure and labels under distribution shifts with three synergistic modules: invariant subgraph identification, environment inference, and invariant learning. DIR~\cite{wuhandling} similarly does not presuppose explicit environmental divisions, but employs a GNN-based mechanism to split the input graph into invariant and variant components, which are then processed by a distribution intervener to minimize invariant risk across varying environments. GSAT~\cite{miao2022interpretable} leverages an attention mechanism enhanced by stochasticity to focus on invariant subgraphs during distribution shifts. Other methods such as EERM~\cite{wuhandling}, DIDA~\cite{zhang2022dynamic}, CIGA~\cite{chen2022learning}, StableGNN~\cite{fan2023generalizing} also explore similar concepts. Despite their progress, these existing methods focus on combining the invariance principle with MPNNs~\cite{guo2024investigating,chen2024does,liu2023flood,li2022ood}, which can not be directly applied to GTs since their architectures are fundamentally different. To our knowledge, we are the first to combine graph invariant learning principles with GTs.

%% file: sections/6-Con.tex
\section{Conclusion}\label{sec:conclusion}
In the paper, we introduce a novel \name~designed to empower Graph Transformer's capable of handling graph data distribution shifts. Three tailored modules, including \disentangler, \encoder, and invariant GT learning, are jointly optimized to enable the graph Transformer to capture invariant relationships between predictive graph structural information and labels. Extensive experiments on synthetic and real-world datasets demonstrate the superiority of \name, along with detailed analyses and ablation studies. In the future, we plan to extend our method for more types of graphs, such as dynamic graphs~\cite{wu2024feasibility,kazemi2020representation} and heterogeneous graphs~\cite{hu2020heterogeneous,zhang2019heterogeneous,wang2019heterogeneous}.

%% file: appendices/Impact.tex
\section{Broader Impacts}\label{sec:impact}
Given the broad applicability of GTs and their susceptibility to distribution shifts and spurious correlations, developing OOD-robust GTs is critical—particularly for real-world applications, where OOD data prevalence is a fundamental challenge. Our work establishes a novel framework to address this problem, aiming to enhance the adaptability and societal impact of GTs. Furthermore, this study adheres to rigorous ethical standards: it involves no human subjects, employs only publicly available datasets, and raises no concerns regarding harmful applications, biases, privacy violations, or legal non-compliance.

%% file: appendices/Limitation.tex
\section{Limitations}\label{sec:limitation}
One limitation of our paper is that we focus on most simple graphs and its worthy extending to other types of graphs, such as dynamic graphs, heterogeneous graphs, and hyper-graphs. Besides, we only verify the effectiveness of our method in widely adopted benchmarks, and we plan to conduct experiments in more real-world applications.

%% file: appendices/Notation.tex
\section{Notations}
\begin{table*}
    \caption{Notations.}
    \label{tab:notation}
    \begin{tabular}{cl} \toprule 
         Notation&Description\\  \midrule  
         $\mathbb{G},\mathbb{Y}$&The graph space and the label space\\
         $\rv{G},\rv{Y}$&The random variables of graph and label\\
         $G,Y$&An instance of graph and label\\
         $\mathcal{G}$&A graph dataset\\
         $\rv{G}_C$&The invariant rationale (causal subgraph) \\
         $\rv{G}_S$&The variant subgraph (complement of $\rv{G}_C$)\\
         $f_Y$&The oracle structural equation model\\
         $\tilde{\mathcal{P}}$&Any distribution over $\mathcal{P}(\rv{G}, \rv{Y})$\\
         $h$&The complete model comprising $h_{\hat{Y}} \circ h_g \circ \mathrm{Softmax} \circ h_l$\\
         $h_l$&Layers preceding the softmax function in the disentangler $\Phi$\\
         $h_g$&Layers following the softmax function in the disentangler $\Phi$\\
         $h_{\hat{Y}}$&The predictor mapping $\rv{G}_C$ to $\rv{Y}$\\
         $\hat{h}$&A function that satisfies the invariant principle\\
         $n_c$&Number of distinct instances of $(i,j)_c$ (node relationships in $\rv{G}_C$)\\
         $n_s$&Number of distinct instances of $(i,j)_s$ (node relationships in $\rv{G}_S$)\\
         $M$&Upper bound of $\hat{h}_l(\rv{G})_{(i,j)_c}$\\
         $m$&Lower bound of $\hat{h}_l(\rv{G})_{(i,j)_s}$\\
         $t$&Temperature parameter\\
         $H$&Entropy\\
         $p_i$&Probability of the $i$-th class\\
         $e_i$&Logit value for the $i$-th class\\
         $\Phi$&The subgraph disentangler\\\bottomrule
    \end{tabular}
\end{table*}
We summarize the key notations and the corresponding descriptions in Table~\ref{tab:notation}.

%% file: appendices/Theory.tex
\section{Theory}\label{sec:theory}
\subsection{Assumption}
\begin{assumption}
There exists a rationale $\rv{G}_C \subseteq \rv{G}$, such that the structural equation model
\begin{displaymath}
    Y\leftarrow f_Y(\rv{G}_C,\epsilon_Y),\epsilon_Y\perp \rv{G}_C
\end{displaymath}
and the probability relation
\begin{displaymath}
    \rv{G}_S\perp \rv{Y} \mid \rv{G}_C
\end{displaymath}
hold for every distribution $\tilde{\mathcal{P}}$ over $\mathcal{P}(\rv{G}, \rv{Y})$, where $\rv{G}_S$ denotes the complement of $\rv{G}_C$. Also, we denote $f_Y$ as the oracle structural equation model.
\label{assumption: invariant rationale}
\end{assumption}

\subsection{Proofs}\label{subsec:proof}

\softmaxdisperse*

\begin{proof}
Following the DIR~\cite{wudiscovering}, we can deduce that the oracle function $f_Y$ from Assumption \ref{assumption: invariant rationale}, which is given 
$\rv{G}_C$, satisfies the principle. Moreover, there exists a unique structure equation model $f_Y$  that adheres to the principle. Consequently, if any model $h$ comprising Eq.~\eqref{eq:complementatt} satisfies the principle, then this model effectively represents the oracle function $f_Y$ given $\rv{G}_C$.

For simplicity, we denote $h=h_{\hat{Y}} \circ h_g \circ \mathrm{Softmax} \circ h_l$, where $h_l$ is the layers preceding the softmax function, and $h_g$ is the layers following the softmax function in the distangler $\Phi$. let's denote $\hat{h} = \hat{h}_{\hat{Y}} \circ \hat{h}_g \circ \mathrm{Softmax} \circ \hat{h}_l$ as a function that satisfies the principle. By definition, $\hat{h}_g(\mathrm{Softmax}(\hat{h}_l(\rv{G}))) = \rv{G}_C$ and $\hat{h}_{\hat{Y}}(\rv{G}_C) = \rv{Y}$. 

Due to two key reasons:
\begin{itemize}[leftmargin=1em]
    \item $\hat{h}_l$ is the sole component responsible for distinguishing $\rv{G}_C$ and $\rv{G}_S$ from $\rv{G}$;
    \item $\hat{h}_l$ includes constraints (e.g., BatchNorm) that bound feature vectors and attention,
\end{itemize}
$\hat{h}_l(\rv{G})$ is guaranteed to stay within a controlled range. Thus, for all nodes $(i,j)$:
\begin{displaymath}
0 < \hat{h}_l(\rv{G})_{(i,j)_c} \leq M, m \leq \hat{h}_l(\rv{G})_{(i,j)_s} < 0,
\end{displaymath}
where $M$ is the upper bound of $\hat{h}_l(\rv{G})_{(i,j)_c}$, $m$ is the lower bound of $\hat{h}_l(\rv{G})_{(i,j)_s}$, $(i,j)_c$ refers to the node relationships in $\rv{G}_C$, and $(i,j)_s$ refers to the node relationships in $\rv{G}_S$.

To prove that for any $i$ and $\epsilon>0$, there must exist a datum $(G,Y)$ such that $(\mathrm{Softmax}( \hat{h}_l(G)))_{(i,j)_c} - (\mathrm{Softmax}(\hat{h}(G)))_{(i,j)_s} < \epsilon$ for all $(i,j)_c$ and $(i,j)_s$, we denote $e_c = \hat{h}_l(G)_{(i,j)_c}$ and $e_s = \hat{h}_l(G)_{(i,j)_s}$. Assume that there are $n_c$ distinct instances of $(i,j)_c$ and $n_s$ distinct instances of $(i,j)_s$. The expression simplifies as follows:
\begin{displaymath}
\begin{split}
&(\mathrm{Softmax}(\hat{h}_l(G)))_{(i,j)_c} - (\mathrm{Softmax}(\hat{h}(G)))_{(i,j)_s} \\
=&\frac{\exp(e_c)}{\sum \exp(e_c) + \sum \exp(e_s)} - \frac{\exp(e_s)}{\sum \exp(e_c) + \sum \exp(e_s)} \\
<&\frac{\exp(M) - \exp(m)}{n_c + n_s \exp(m)}
\end{split}
\end{displaymath}
For all $n_c$ and $n_s$, if $n_c + n_s \exp(m) > \frac{\exp(M) - \exp(m)}{\epsilon}$, then 
\begin{displaymath}
(\mathrm{Softmax}(\hat{h}_l(G)))_{(i,j)_c} - (\mathrm{Softmax}(\hat{h}(G)))_{(i,j)_s} < \epsilon.
\end{displaymath}

Considering the influence of noise, we assume that only if \\$(\mathrm{Softmax}(\hat{h}_l(G)))_{(i,j)_c} - (\mathrm{Softmax}(\hat{h}(G)))_{(i,j)_s} > \epsilon'$, can $\hat{h}_g$ potentially generate the correct $G_c$. Therefore, as the values of $n_c$ and $n_s$ increase such that $n_c + n_s \exp(m) > \frac{\exp(M) - \exp(m)}{\epsilon'}$,
\begin{displaymath}
(\mathrm{Softmax}(\hat{h}_l(G)))_{(i,j)_c} - (\mathrm{Softmax}(\hat{h}(G)))_{(i,j)_s} < \epsilon'.
\end{displaymath}
This implies that $\hat{h}_g(\mathrm{Softmax}( \hat{h}_l(G))) \neq G_c$, leading to a contradiction with $\hat{h}_g(\mathrm{Softmax}(\hat{h}_l(\rv{G}))) = \rv{G}_C$, thereby completing the proof.
\end{proof}

This theory suggests that for different $G$, the softmax-based subgraph disentangler $h_g \circ \mathrm{Softmax} \circ h_l $ may not consistently identify the invariant subgraph $ G_c $. Previous discussions focused on the increase in $ n_c $ and $ n_s $, representing a growth in subgraph size. Additionally, in practical scenarios, due to the influence of noise and variations in the graph data distribution, the gap between $ h_l(G)_{(i,j)_c} $ and $ h_l(G)_{(i,j)_s} $ might be insufficiently distinct. This can cause the subgraph disentangler to confuse invariant and variant subgraphs, particularly with out-of-distribution test data.

\temperature*

\begin{proof}
We begin by expressing the entropy $H$ as follows:
\begin{displaymath}
\begin{split}
    H&=-\sum_{i}p_i\log{p_i} \\
     &=-\sum_{i}p_i(\frac{e_i}{t}-\log{\sum_{k}\mathrm{exp}(\frac{e_k}{t})}) \\
     &=\log{\sum_{k}\mathrm{exp}(\frac{e_k}{t})}-\sum_{i}\frac{p_{i}e_i}{t}
\end{split}
\end{displaymath}
To determine the monotonicity of $H$ with respect to $t$, we take the derivative of this expression with respect to $t$:
\begin{displaymath}
\begin{split}
    \frac{dH}{dt}&=\frac{\sum_{k}-\frac{e_k}{t^2}\rm{exp(\frac{e_k}{t})}}{\sum_{k}\rm{exp}(\frac{e_k}{t})}
                        + \sum_{i}\frac{e_i}{t^2}p_i
                        - \sum_{i}\frac{e_i}{t}\frac{dp_i}{dt}\\
                      &=- \sum_{k}\frac{e_k}{t^2}p_k
                        + \sum_{i}\frac{e_i}{t^2}p_i
                        - \sum_{i}\frac{e_i}{t}[p_i(1-p_i)(-\frac{e_i}{t^2})]\\
                      &=\sum_{i}\frac{{e_{i}}^2}{t^3}p_i(1-p_i)
\end{split}
\end{displaymath}
Since $t > 0$, and $0<p_i<1$, it follows that $\frac{dH}{dt}>0$. Therefore, as the temperature $t$ decreases, the value of $H$ must decrease monotonically.
\end{proof}

%% file: appendices/ExpDetail.tex
\section{Experimental Details}\label{sec:expdetails}

\subsection{Datasets}\label{sec:dataset-detail}
We adopt five graph classification datasets to verify the effectiveness of \name. 

\begin{table}
    \caption{The graph classification accuracy (\%) on testing sets of SP-Motif.}
    \label{tab:shiftlevel}
    \begin{tabular}{lccc} \toprule 
         Methods&b=0.5&b=0.7&b=0.9\\  \midrule  
         GraphGPS&46.89\scalebox{0.75}{±7.56}&52.65\scalebox{0.75}{±7.83}&48.58\scalebox{0.75}{±4.20}\\
         EXPHORMER&40.17\scalebox{0.75}{±6.07}&47.98\scalebox{0.75}{±4.39}&60.35\scalebox{0.75}{±5.80}\\
         \name&\best{82.07}\scalebox{0.75}{±4.71}&\best{82.57}\scalebox{0.75}{±5.57}&\best{72.55}\scalebox{0.75}{±10.43}\\\bottomrule
    \end{tabular}
\end{table}

\begin{figure*}
        \begin{subfigure}{0.33\linewidth}
		\centering
		\includegraphics[width=\linewidth]{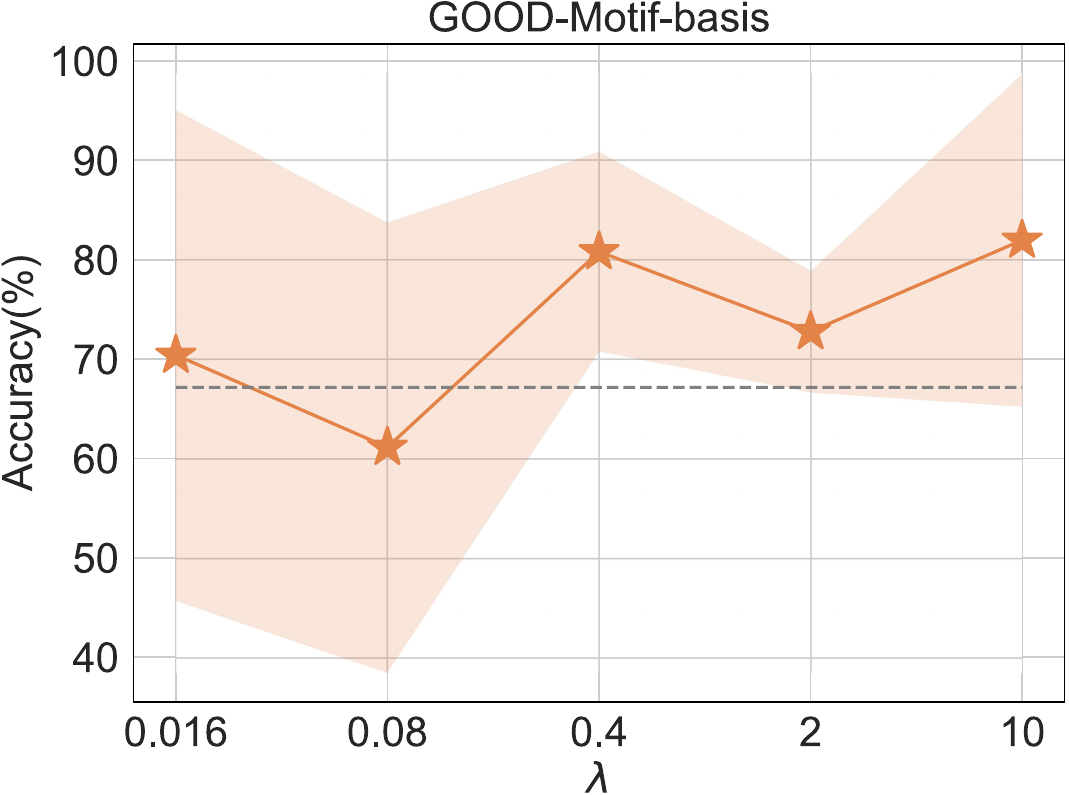}
	\end{subfigure}
        \begin{subfigure}{0.33\linewidth}
		\centering
		\includegraphics[width=\linewidth]{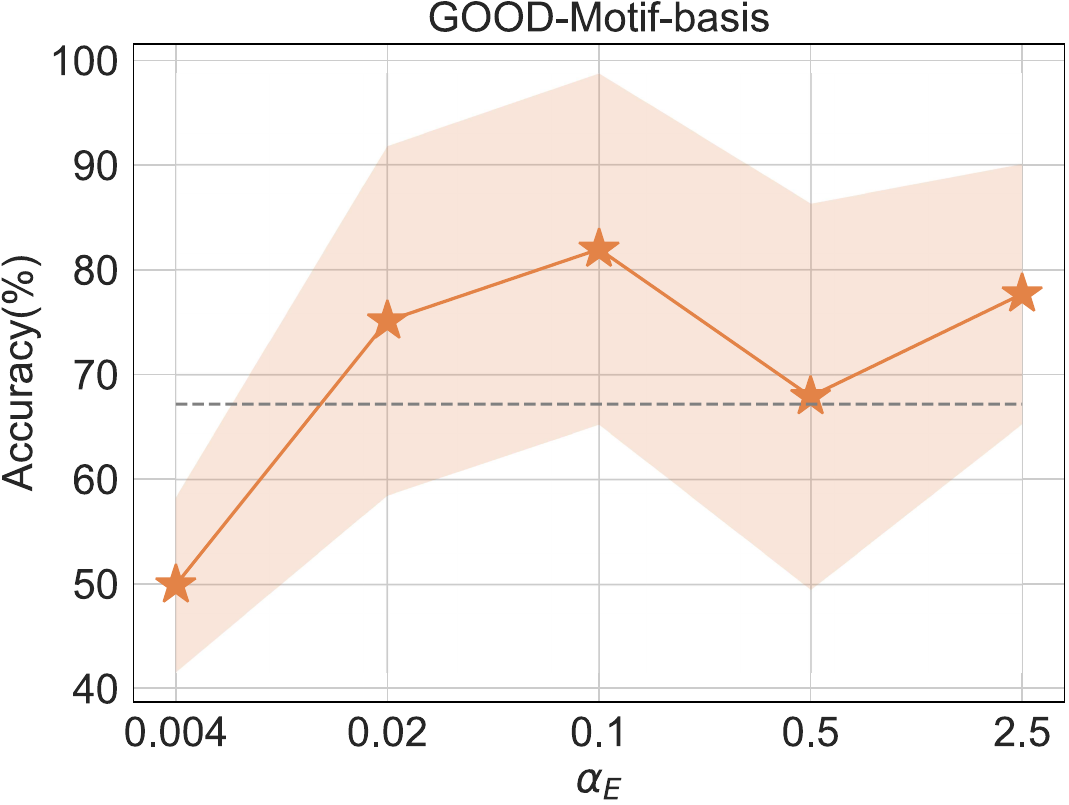}
	\end{subfigure}
        \begin{subfigure}{0.33\linewidth}
		\centering
		\includegraphics[width=\linewidth]{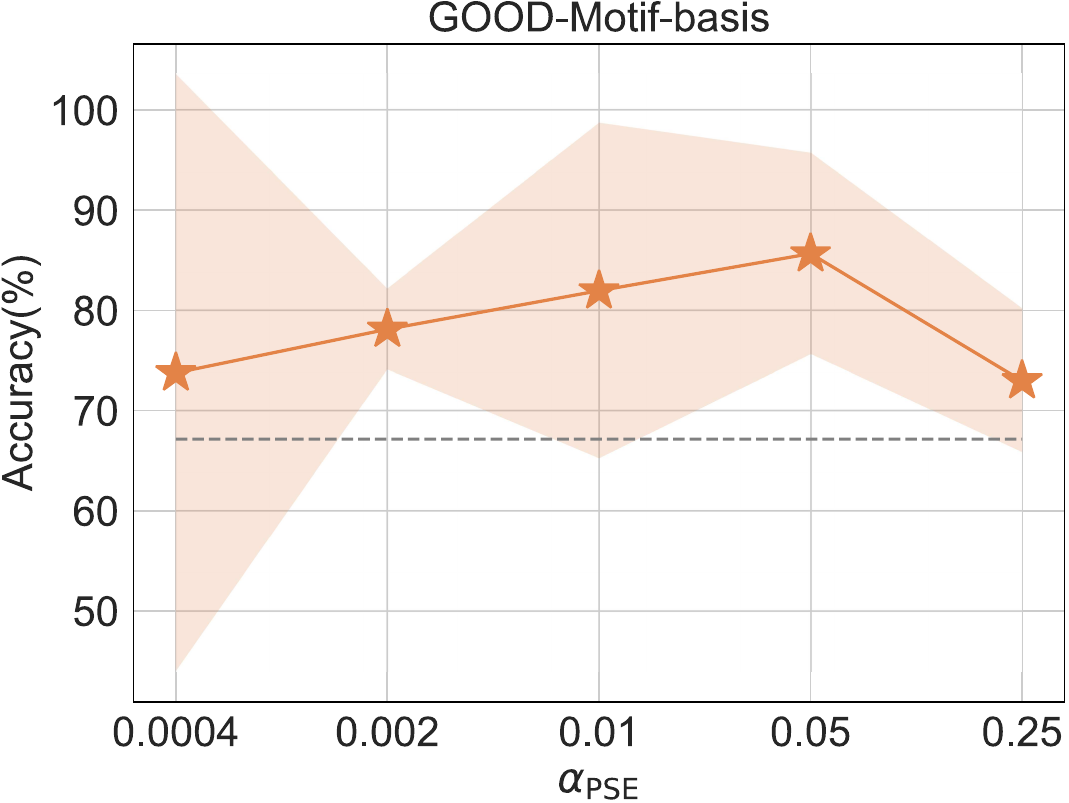}
	\end{subfigure}
	\begin{subfigure}{0.33\linewidth}
		\centering
		\includegraphics[width=\linewidth]{figures/GOOD-Motif-size-I.pdf}
	\end{subfigure}
	\begin{subfigure}{0.33\linewidth}
		\centering
		\includegraphics[width=\linewidth]{figures/GOOD-Motif-size-E.pdf}
	\end{subfigure}
	\begin{subfigure}{0.33\linewidth}
		\centering
		\includegraphics[width=\linewidth]{figures/GOOD-Motif-size-PSE.pdf}
	\end{subfigure}
    \caption{The impact of different hyper-parameters. Orange and red lines denote the results of \name, while grey dashed lines indicate: (1) CIGA's performance on the basis split and (2) EXPHORMER's performance on the size split.}
    \label{fig:hyper-parameter sensitivity 2}
\end{figure*}

GOOD-Motif is a synthetic dataset designed to test OOD algorithms for feature and structure shifts, inspired by the Spurious-Motif dataset~\cite{wudiscovering}. It comprises graphs generated by combining a label-irrelevant base graph (such as a wheel, tree, ladder, star, or path) with a label-determining motif (like a house, cycle, or crane). The base graph type and size can be designated as domain features.

GOOD-HIV is a real-world molecular dataset derived from MoleculeNet~\cite{wu2018moleculenet}, featuring molecular graphs where nodes represent atoms and edges symbolize chemical bonds. The primary task is to predict the molecule's ability to inhibit HIV replication. The dataset is divided based on two domain features: the Bemis-Murcko scaffold, which represents the two-dimensional structural base of the molecule, and the number of nodes, indicating the size of the molecular graph.

GOOD-SST2 is a real-world natural language sentiment analysis dataset adapted from ~\cite{yuan2022explainability}, where each sentence is converted into a grammar tree graph with nodes representing words, and word embeddings serving as node features. It is designed for a binary classification task to determine the sentiment polarity of sentences. Sentence length is chosen as the domain feature, under the assumption that the length should not influence the sentiment polarity.

GOOD-Twitter is a real-world natural language sentiment analysis dataset adapted from ~\cite{yuan2022explainability}, using a transformation similar to SST2 where sentences are turned into grammar tree graphs, with nodes as words and word embeddings as node features. Unlike SST2's binary classification, this dataset involves a three-fold classification task to predict one of three sentiment polarities for each sentence. Sentence lengths are chosen as the domains.

DrugOOD~\cite{ji2023drugood} is employed in the task of ligand-based affinity prediction (LBAP), with the core noise level and IC50 measurement type as the domain features. We only use a dataset of it, named LBAP-core-ic50.

\subsection{Baselines}\label{sec:baseline-detail}
We provide detailed descriptions of baselines as follows:
\begin{itemize}[leftmargin=1em]
    \item ERM: is trained with the standard empirical risk minimizing.
    \item IRM~\cite{arjovsky2019invariant}: features a regularized objective that mandates the uniform optimality of a single classifier across various environments. This method exemplifies invariant learning by aiming to discover data representations or features where the optimal predictor remains consistent across all settings.
    \item VREx~\cite{krueger2021out}: adopts a variance risk extrapolation approach that promotes minimizing training risks while enhancing the consistency of these risks across different training conditions. This strategy has been demonstrated to uncover the causal mechanisms underlying target variables and offers resilience against distribution shifts. 
    \item DIR~\cite{wudiscovering}: applies interventions to graph structures, generating interventional distributions that enhance the model's ability to generalize across different scenarios.
    \item GREA~\cite{liu2022graph}: proposes an efficient graph rationalization framework that leverages environment-based augmentations to create virtual training examples. By performing all operations in a latent space, it avoids the computational overhead of explicit graph decoding and encoding while improving rationale identification accuracy.
    \item GIL~\cite{li2022learning}: addresses graph-level OOD generalization by learning invariant relationships between predictive subgraphs and labels under distribution shifts. Since environment labels are typically unavailable, GIL proposes a framework with three co-optimized modules.
    \item GSAT~\cite{miao2022interpretable}: focuses on constructing GNNs that are not only inherently interpretable but also more generalizable by imposing penalties on the excessive use of information from the input data.
    \item CIGA~\cite{chen2022learning}: establishes a framework that distinguishes between the causal and non-causal components of graph data, specifically identifying how these components contribute to invariant features essential for prediction tasks. The framework also offers a comprehensive theoretical discussion on graph-based OOD generalization.
    \item iMoLD~\cite{zhuang2023learning}: proposes a novel framework for learning invariant molecular representations against distribution shifts through a "first-encoding-then-separation" strategy.
    \item CAL+~\cite{sui2024enhancing}: introduces a causal learning framework that addresses OOD generalization in graph classification by mitigating the confounding effect of shortcut features. It leverages backdoor adjustment through a memory bank to combine causal features with diverse shortcuts, while using prototypes to ensure intra-class causal consistency. 
    \item Graphormer~\cite{ying2021transformers}: introduces a novel approach by utilizing pairwise graph or 3D distances to craft relative positional encodings.
    \item GraphGPS~\cite{rampavsek2022recipe}: offers a versatile framework that integrates message-passing networks with attention mechanisms. It supports a variety of embeddings, both positional and structural, and accommodates sparse transformer models, enhancing flexibility and adaptability in graph neural networks.
    \item EXPHORMER~\cite{shirzad2023exphormer}: represents a breakthrough in sparse graph transformer architectures by introducing two innovative sparse network designs utilizing virtual nodes and expander graphs.
\end{itemize}

%% file: appendices/MoreExp.tex
\section{Additional Experimental Results}

\subsection{Hyper-parameter Sensitivity Analysis}\label{subsec:hyper-parameter}

We further evaluate the sensitivity of hyper-parameters, including the invariant learning hyper-parameter $\lambda$, the entropy loss hyper-parameter $\alpha_E$ and the encoding loss hyper-parameter $\alpha_{\mathrm{PSE}}$ on the basis and size splits of GOOD-Motif in Figure \ref{fig:hyper-parameter sensitivity 2}. Our method is not very sensitive to hyper-parameters and outperforms both the general OOD generalization method CIGA and graph Transformer EXPHORMER with a wide range of hyper-parameters choices. Besides, many of our hyper-parameters have concrete meanings and do not necessitate a resource-intensive tuning:
\begin{itemize}[leftmargin=1em]
    \item Regarding $\alpha_S$: This parameter solely controls the influence of the classification loss for the classifier $w_{\tilde{s}}(\cdot)$. Actually, in our experiments, $\alpha_S$ is consistently set to 1.
    \item Regarding $\alpha_E,\alpha_{\mathrm{PSE}}$: They can be relatively straightforwardly determined based on the relative magnitudes of each loss component. Specifically, we can let $\alpha_E\mathcal{L}_E$ and $\alpha_{\mathrm{PSE}}\mathcal{L}_{\mathrm{PSE}}$ to be of comparable scale, while ensuring $\mathcal{L}_I$ dominates both.
\end{itemize}
In sum, the hyper-parameters of our method can be practically tuned without affecting its applicability. 

\subsection{Running Time Analysis}\label{subsec:runningtime}

\begin{table*}
\centering
    \caption{Averaged training and inference time (sec.) per epoch of various methods on GOOD-Motif.}
    \label{tab:runningtime}
    \begin{tabular}{lcccc} \toprule 
         Methods&Graphormer&GraphGPS&EXPHORMER&\name\\  \midrule  
         Training&7.7&9.5&11.9&28.6\\
         Inference&2.1&1.7&5.6&7.1\\
         OOD Performance&40.21\scalebox{0.75}{±1.55}&46.48\scalebox{0.75}{±1.80}&46.44\scalebox{0.75}{±11.81}&81.97\scalebox{0.75}{±16.74}\\\bottomrule
    \end{tabular}
\end{table*}

To evaluate the computational overhead introduced by \name's architecture and auxiliary objectives, we conduct a comparative analysis of average training and inference times across different GTs on the basis split of GOOD-Motif benchmark. Key factors influencing runtime—including batch size and hardware configuration (Tesla V100S-PCIE-32GB GPUs, Linux server with 16-core Intel Xeon Gold 5218 CPUs @ 2.30GHz, and Ubuntu 22.04 LTS)—are held constant during testing. As summarized in Table~\ref{tab:runningtime}, the results align with our time complexity analysis in Section~\ref{subsec:discussion}, demonstrating that \name~maintains computational efficiency on par with other GTs. The marginal overhead stems primarily from invariant learning and PSEs, which are justified by their significant performance gains.

\subsection{Shift Level Insights}\label{subsec:shift-level-insights}

To evaluate \name's OOD generalization ability in different shift levels, we add experiments with varying shift severity (controlled by b) on the SP-Motif Dataset~\cite{wudiscovering, li2022learning}. The results are shown in Table~\ref{tab:shiftlevel}. Experimentally, \name~performs strongly under moderate distribution shifts, though some degradation occurs under extreme shifts, mainly due to decreased accuracy in invariant subgraph discovery. Performance may also be limited in more extreme scenarios-such as semantic shifts in OOD detection~\cite{cai2025out} with unseen invariant subgraphs. That said, we wish to emphasize that \name~is specifically designed for OOD generalization.

\subsection{Graph Transformer vs. MLP as Predictor}\label{subsec:GT-vs-MLP}

We employ GTs as predictors instead of MLPs to enhance node representations with explicit structural information. To validate whether GTs indeed outperform standard MLPs in this role, we conduct a comparative ablation study, substituting the GT in Eq.~\eqref{eq:graphrepresentation} with an MLP, with results summarized in Table~\ref{tab:GT-vs-MLP}. Our analysis confirms that GTs consistently achieve superior performance compared to general MLPs.

\begin{table}
    \caption{Performance comparison between GT and MLP predictors.}
    \label{tab:GT-vs-MLP}
    \begin{tabular}{lcccc} \toprule 
         &\multicolumn{2}{c}{GOOD-Motif}&\multicolumn{2}{c}{GOOD-HIV}\\ \cline{2-5} 
         &basis&size&scaffold&size\\ \midrule  
         MLP&70.66\scalebox{0.75}{±9.36}&79.36\scalebox{0.75}{±5.16}&68.34\scalebox{0.75}{±4.43}&59.70\scalebox{0.75}{±3.06}\\
         GT&\best{81.97}\scalebox{0.75}{±16.74}&\best{80.02}\scalebox{0.75}{±7.94}&\best{70.38}\scalebox{0.75}{±0.27}&\best{62.69}\scalebox{0.75}{±0.11}\\
        \bottomrule
    \end{tabular}
\end{table}